\documentclass{article} 
\usepackage{iclr2026_conference,times}

\usepackage{amsmath,amsfonts,bm}

\def\eqref#1{equation~\ref{#1}}

\def\1{\bm{1}}

\DeclareMathAlphabet{\mathsfit}{\encodingdefault}{\sfdefault}{m}{sl}
\SetMathAlphabet{\mathsfit}{bold}{\encodingdefault}{\sfdefault}{bx}{n}

\usepackage{xcolor}
\definecolor{mycitecolor}{HTML}{3498DC}
\definecolor{mylinkcolor}{HTML}{E74D3B}
\definecolor{myblue}{HTML}{4594c1}
\definecolor{myurlcolor}{HTML}{980000}
\usepackage{hyperref}
\usepackage{url}
\usepackage{graphicx}
\usepackage{lipsum}
\usepackage{wrapfig}
\usepackage{microtype}

\usepackage{inconsolata}

\usepackage{graphicx}
\usepackage{amsmath}
\usepackage{amsfonts}
\usepackage{bm}
\usepackage{array}
\usepackage{booktabs}
\usepackage{multirow}
\usepackage{tabularx}
\usepackage{ulem}
\usepackage{float}
\usepackage{subfigure}
\usepackage{hyperref}
\usepackage[most]{tcolorbox}
\usepackage{longtable}
\usepackage{colortbl}
\usepackage{arydshln}
\usepackage{makecell}
\usepackage{placeins}
\usepackage{titletoc}
\definecolor{mylinkcolor}{HTML}{E74D3B}

\usepackage{soul}       
\usepackage{xcolor}
\usepackage{xurl}       
\definecolor{blue}{RGB}{33,150,243}

\tcbset{
  mytrace/.style={
    enhanced,
    breakable,
    colback=gray!2,
    colframe=gray!35,
    boxrule=0.3pt,
    arc=1pt,
    left=4pt,right=4pt,top=3pt,bottom=3pt,
    fontupper=\small,
    before skip=4pt, after skip=6pt
  }
}
\usepackage{fvextra} 
\DefineVerbatimEnvironment{jsoncode}{Verbatim}{
  breaklines,          
  breakanywhere,       
  breaksymbolleft=\tiny\color{gray}{$\hookrightarrow$}, 
  breaksymbolindent=0pt, breaksymbolsep=0pt,
  fontsize=\small
}

\title{UltraHorizon: Benchmarking Agent Capabilities in Ultra Long-Horizon Scenarios}
\newcommand{\benchname}{\textit{\textbf{\textrm{UltraHorizon}}}}

\author{
\textbf{Haotian Luo}\textsuperscript{1,2,*} \quad
\textbf{Huaisong Zhang}\textsuperscript{1,3,*} \quad
\textbf{Xuelin Zhang}\textsuperscript{1,2,*} \quad
\textbf{Haoyu Wang}\textsuperscript{3,*} \\
\textbf{Zeyu Qin}\textsuperscript{4,*} \quad
\textbf{Wenjie Lu}\textsuperscript{1,*} \quad
\textbf{Guozheng Ma}\textsuperscript{5} \quad
\textbf{Haiying He}\textsuperscript{6} \quad
\textbf{Yingsha Xie}\textsuperscript{2} \\
\textbf{Qiyang Zhou}\textsuperscript{2} \quad
\textbf{Zixuan Hu}\textsuperscript{5} \quad
\textbf{Hongze Mi}\textsuperscript{7} \quad
\textbf{Yibo Wang}\textsuperscript{3} \quad
\textbf{Naiqiang Tan}\textsuperscript{1,\dag,\ddag} \\
\textbf{Hong Chen}\textsuperscript{8} \quad
\textbf{Yi R. Fung}\textsuperscript{4} \quad
\textbf{Chun Yuan}\textsuperscript{3,\dag} \quad
\textbf{Li Shen}\textsuperscript{2,\dag} \\
\\
\textsuperscript{1} Didichuxing Co. Ltd \quad
\textsuperscript{2} Sun Yat-sen University \quad
\textsuperscript{3} Tsinghua University \quad
\textsuperscript{4} HKUST \\
\textsuperscript{5} Nanyang Technological University \quad
\textsuperscript{6} China Agricultural University \\
\textsuperscript{7} Tianjin University \quad
\textsuperscript{8} Huazhong Agricultural University \\
\\
\textsuperscript{*} Equal contribution \quad
\textsuperscript{\dag} Corresponding author \quad
\textsuperscript{\ddag} Project leader
}

\iclrfinalcopy 
\begin{document}

\maketitle

\vspace{-6mm}
\begin{abstract}

Autonomous agents have recently achieved remarkable progress across diverse domains, yet most evaluations focus on short-horizon, fully observable tasks. In contrast, many critical real-world tasks, such as large-scale software development, commercial investment, and scientific discovery, unfold in long-horizon and partially observable scenarios where success hinges on sustained reasoning, planning, memory management, and tool use. Existing benchmarks rarely capture these long-horizon challenges, leaving a gap in systematic evaluation. To bridge this gap, we introduce \benchname, a novel benchmark that measures the foundational capabilities essential for complex real-world challenges. We use exploration as a unifying task across three distinct environments to validate these core competencies. Agents are designed in long-horizon discovery tasks where they must iteratively uncover hidden rules through sustained reasoning, planning, memory and tools management, and interaction with environments. Under the heaviest scale setting, trajectories average \textbf{200k+} tokens and \textbf{400+} tool calls, whereas in standard configurations they still exceed \textbf{35k} tokens and involve more than \textbf{60} tool calls on average. Our extensive experiments reveal that LLM-agents consistently underperform in these settings, whereas human participants achieve higher scores, underscoring a persistent gap in agents' long-horizon abilities. We also observe that simple scaling fails in our task. To better illustrate the failure of agents, we conduct an in-depth analysis of collected trajectories. We identify eight types of errors and attribute them to two primary causes: in-context locking and functional fundamental capability gaps. \href{https://github.com/StarDewXXX/UltraHorizon}{Our code will be available here.}
\end{abstract}

\begin{figure}[htbp]
\centering
\includegraphics[width=1.0\textwidth]{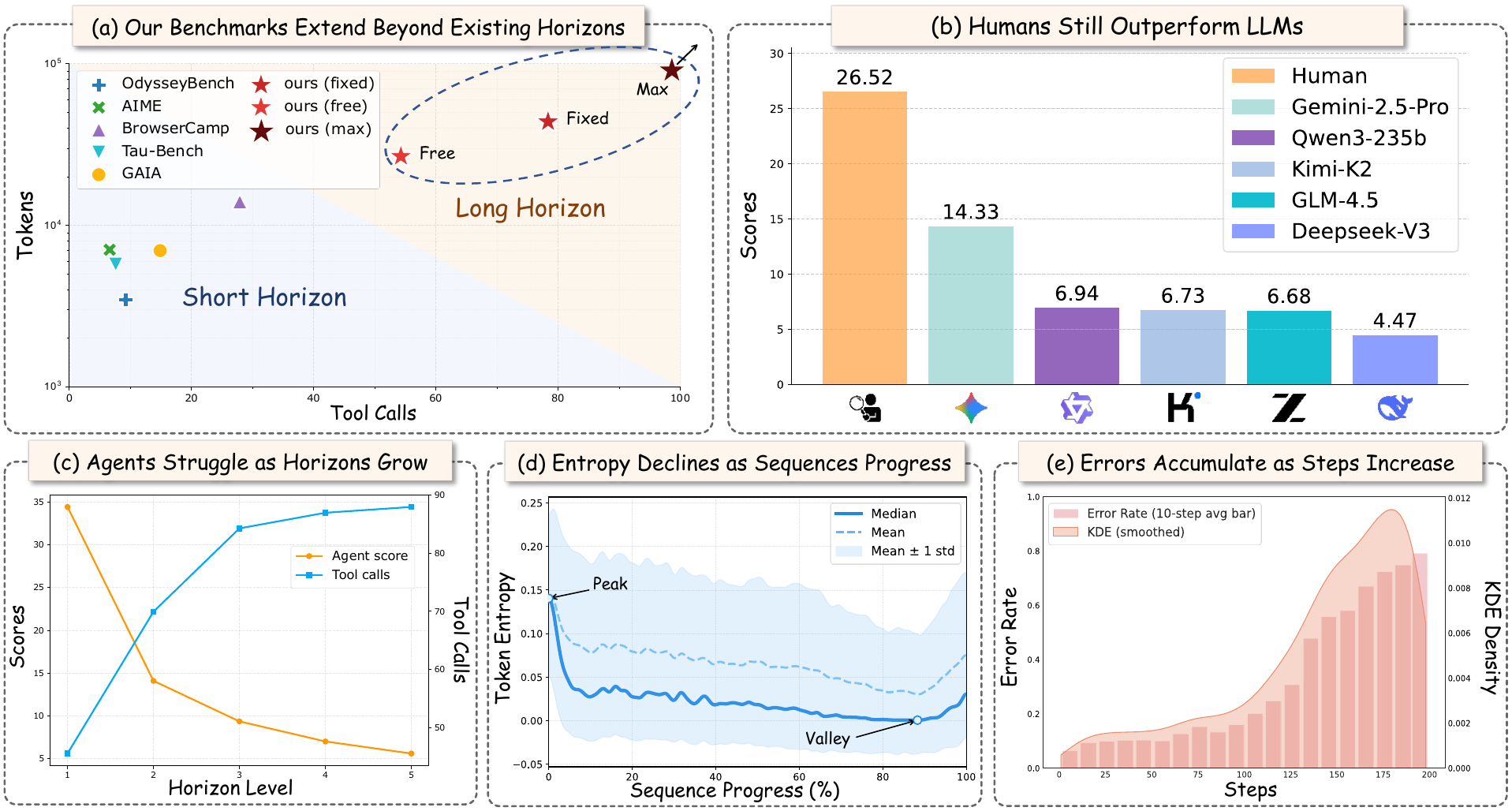}
\caption{Overview of \benchname{} and agent performance dynamics across different horizons. 
(a) Our benchmarks occupy the long-horizon regime, extending beyond existing suites. 
(b) Human participants still outperform leading LLMs on \benchname{}. 
(c) Agent scores decline while tool calls increase as horizon level grows. 
(d) Token entropy declines as sequences progress, with a slight late-stage uptick. 
(e) Error rates accumulate with more interaction steps.}
\label{fig:highlight}
\end{figure}


\section{Introduction}
\label{sec:intro}
Autonomous language agents have made rapid progress in planning, tool use, and interactive decision-making across various fields. 
LLM-based agents demonstrate superiority and proficiency in multi-tool utilization \citep{zhang2025toolexpnet, huang2024planning}, long-context reasoning \citep{an2024make, wang2025odysseybench}, multi-step planning \citep{rawat2025pre,lin2025se}, and interactive environmental engagement \citep{chen2024automanual, ma2025caution}. 

To systematically evaluate these capabilities, numerous agent benchmarks have been proposed. For instance, $\tau$-Bench \citep{yao2024tau} focuses on multi-turn interactions with simulated users, SWE-Bench \citep{zhang2025swe} tasks models with fixing real GitHub issues, AgentBench \citep{liu2023agentbench} evaluates decision-making in diverse simulated environments, and ToolExpNet \citep{zhang2025toolexpnet} assesses multi-tool selection strategies. 

While these benchmarks have advanced our understanding of agent performance, they generally remain short-horizon. As shown in Figure \ref{fig:highlight}(a), typical benchmarks involve only a few thousand tokens and fewer than 30 tool calls per trajectory. For example, OdysseyBench \citep{wang2025odysseybench} averages 3.5k tokens and 9 tool calls, and Tau-Bench 5.7k tokens and 8 tool calls. This highlights that existing benchmarks capture relatively lightweight interactions, but do not sufficiently stress-test the sustained reasoning, planning, and memory management essential for long-horizon challenges.

Moreover, most existing benchmarks operate in fully observable or quasi-deterministic settings, where the environment is transparent and the relevant task is defined by immediate objectives. They rarely capture the challenges posed by partially observable environments, where agents must iteratively form hypotheses, manage uncertainty, and conduct systematic exploration before reaching a solution. This leaves a critical gap in evaluating the long-horizon reasoning and exploration capacities essential for real-world applications.

In contrast, critical real-world tasks, such as large-scale software development \citep{jahic2024state,wei2024requirements,jin2024llms,zhang2025knowledge}, commercial investment \citep{mahdavi2025integrating,li2025can}, and scientific discovery \citep{swanson2025virtual,wang2025geneagent,qu2025crispr}, unfold over long horizons and in partially observable settings.

To bridge this gap, we introduce \benchname, a new benchmark that measures foundational skills for long-horizon, partially observable task solving. \benchname\ uses exploration as a unifying task format across three distinct environments, each designed to probe complementary competencies in sustained reasoning, planning, memory management, and tool use. Agents are placed in discovery-oriented tasks where they must uncover and exploit hidden rules through iterative interaction. Trajectories average \textbf{200k+} tokens and \textbf{400+} tool calls in the heaviest setting, while in standard configurations they still exceed \textbf{35k} tokens and involve more than \textbf{60} tool calls on average, underscoring the benchmark’s extended horizons and interaction complexity.

We conduct extensive experiments on \benchname\ with state-of-the-art LLMs. Despite strong performance on short-horizon tasks, these agents consistently underperform, while human participants achieve higher scores. Analyzing agent trajectories, we identify nine recurring error patterns rooted in two primary causes: (i) \textbf{In-context Locking}, where agents become locked on initial patterns, and (ii) \textbf{Foundational Capability Gaps}, reflecting deficiencies in long-horizon reasoning, memory use, and tool calling.

In summary, our work makes the following contributions:
\begin{itemize}
    \item We introduce \benchname, a novel benchmark explicitly designed to evaluate LLM-based agents in \textbf{long-horizon, partially observable} environments. Unlike prior benchmarks that emphasize short trajectories and fully observable settings, \benchname\ requires agents to perform sustained reasoning, planning, memory management, and tool calling.
    
    \item Through extensive experiments, we show that LLM-Agents consistently underperform on \benchname. In contrast, human participants demonstrate much higher success rates, highlighting a significant capability gap in long-horizon tasks. We also observe that simple scaling fails due to context overload and propose a simple yet effective scaling strategy.
    
    \item By analyzing collected agent trajectories, we propose a two-level failure categorization framework that distinguishes root causes from observable manifestations with a highlight on \textbf{in-context locking}.

\end{itemize}


\section{Related Work}
\label{sec:related}

\textbf{Benchmarks for language agents and tool use.}
Recently, several benchmarks have been introduced to evaluate the tool-use capabilities of LLM-based agents across various domains. For example, $\tau$-Bench \citep{yao2024tau} focuses on multi-turn interactions with simulated users, which are conducted with short-horizon interactions and fail to assess agents' capabilities in extended exploration scenarios adequately. SWE-Bench \cite{zhang2025swe} tasks models with fixing real GitHub issues and verifies the solutions using automated tests. Meanwhile, BrowseComp \citep{wei2025browsecomp} and GAIA \citep{mialon2023gaia} challenge agents with difficult questions that require persistent multi-hop search and reasoning, web browsing, and API usage. Others like AgentBench \citep{liu2023agentbench} evaluate decision-making in diverse simulated environments with multi-turn challenges. 
ZeroSumEval \citep{khan2025zerosumeval} conducts arena-style evaluations on LLMs in classic strategy games like chess and poker, which are limited to known rules and target the success or highest scores, ignoring the investigation of complex and unknown rules. 
In short, these benchmarks evaluate agents in specific environments with a focus on shorter-horizon tasks with known rules, which emphasizes the need for assessing long-horizon exploration and rule discovery in partially observable settings.


\textbf{Long-context reasoning evaluation.} Long-context reasoning evaluates an agent’s ability to process and reason over substantial inputs. Recent studies show that even advanced LLMs often fail when required to extract and chain information from lengthy documents. For instance, \cite{kuratov2024babilong} introduces the BABILong benchmark, consisting of reasoning tasks where the necessary facts are scattered across long texts. Similarly, \cite{ling2025longreason} proposes LongReason, a synthetic benchmark with hundreds of questions embedded in expanded contexts. A widely used related test is ``needle-in-a-haystack" (NIAH). In this setup, a specific key fact is hidden within a large block of background text, and the model must retrieve and utilize it. RULER \citep{hsieh2024ruler} extends the NIAH test by incorporating varied needle types and counts, and introduces multi-hop tracing and aggregation tasks to evaluate behaviors beyond context-based search. \benchname{} differs from existing long-context reasoning benchmarks in that it emphasizes interactive, exploratory tasks rather than passive comprehension or question-answering over a static context.

\section{UltraHorizon}
\label{sec:benchmark}
\begin{figure}[htbp]
\centering
\includegraphics[width=\linewidth]{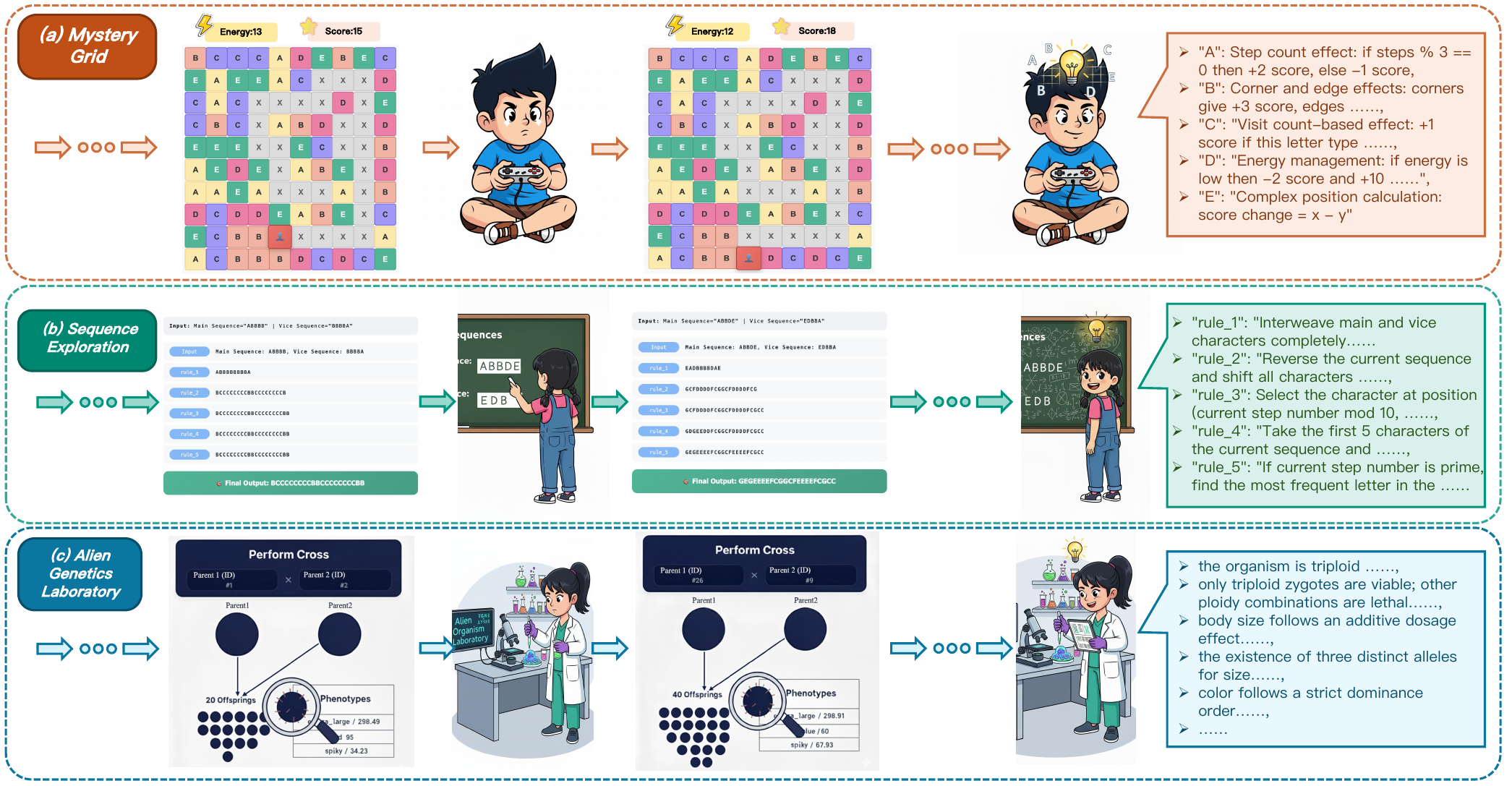}
\caption{The illustration of three distinct environments in \benchname{}.
\textbf{Mystery Grid}: agents explore a $10\times10$ grid under energy and step budgets to infer the hidden mapping from symbols A--E to score/energy effects;
\textbf{Sequence Exploration}: agents design paired length-5 sequences and, from deterministic outputs of a fixed five-rule pipeline, induce the underlying transformation rules;
\textbf{Alien Genetics Laboratory}: agents conduct controlled crosses in a triploid organism to recover inheritance mechanisms, including gamete formation, dosage/dominance patterns, and lethal combinations. In Appendix \ref{app.html}, we provide a screenshot of an intuitive GUI-based version of the environment.} 
\label{fig:pipline}
\end{figure}

\subsection{Design Principles}

In designing this benchmark, we adhere to several principles to ensure that it effectively evaluates the agents' overall abilities in long-horizon and partially observable scenarios:

\begin{itemize}
\item \textbf{Temporal Depth}
The benchmark requires agents to cope with long interaction sequences where outcomes depend on delayed consequences and extended dependencies, highlighting challenges unique to deep time horizons.

\item \textbf{Consistency} The design of this benchmark follows a rigorous set of rules and logical structure to ensure internal coherence. The rules of the environment are not dependent on external knowledge, and all factors influencing the environment can be discovered and reasoned about through agent-environment interactions.

\item \textbf{Unknowability} The benchmark intentionally includes unknown elements and rules that agents must uncover through exploration. Success depends on generating hypotheses, discovering rules, and refining understanding over time, as agents begin without full knowledge of the environment’s dynamics.

\item \textbf{Real-world Ability Relevance}
Tasks mirror real-world demands by requiring sustained reasoning, planning, decision-making, memory management, and adaptation in dynamic, partially observable environments.

\end{itemize}

\subsection{Environments}
Based on the aforementioned design principles, we have carefully crafted three distinct environments, which are illustrated in Figure \ref{fig:pipline}. Due to space limitations, we present only an overview of the environments here. For more detailed information, including prompts and a list of tools, please refer to the Appendices \ref{sec:a1} to \ref{sec:a2}.

\subsubsection{Environment I: Mystery Grid}

The \textit{Mystery Grid Environment} is a $10 \times 10$ grid world. Five symbols (A-E) each encode hidden, fixed effects on score or energy that remain consistent across episodes. Agents start with limited energy, incur a cost per move, and can explore up to $30$ steps per episode and $N$ steps in one game run. Multiple resets are permitted, enabling iterative hypothesis testing. The letter rules range from simple effects to conditional dependencies on factors such as energy or position. At the end of exploration, the agent must submit an explicit mapping of symbol-effect relations, which is evaluated for correctness and precision.


\subsubsection{Environment II: Sequence Exploration}

The \textit{Sequence Exploration Environment} evaluates an agent’s capacity for systematic experimentation and hidden-rule inference in symbolic domains. In each interaction, the agent provides two sequences of length five over the alphabet A–E, subject to constraints ensuring diversity of symbols. The environment then applies a fixed chain of five latent transformation rules to the input, producing intermediate and final outputs. These rules remain deterministic and invariant across trials but are concealed from the agent. After $N$ exploratory steps, the agent must provide a complete natural language description of the five transformation rules.


\subsubsection{Environment III: Alien Genetics Laboratory}

The \textit{Alien Genetics Laboratory} models a triploid extraterrestrial organism whose inheritance rules may diverge from terrestrial biology. Agents interact with the environment through controlled genetic experiments, including hybridization, organism queries, and trait analysis. Each organism possesses three genomic sets, and trait expression may involve additive effects, dominance hierarchies, or lethal genotype combinations. The evaluation requires agents to articulate a complete model of the underlying genetic system after $N$ experiments, covering gamete formation, trait inheritance patterns, and lethal interactions.




\section{Experiments}
\label{sec:experiment}

\subsection{Experimental Setup}

\textbf{Models and Agent Settings.} To systematically evaluate the long-horizon exploration, memory management, and rule-investigation capabilities of LLM agents, we conduct extensive experiments on \benchname{} with five widely-used LLMs covering both proprietary and open-source families, including Gemini-2.5-Pro \citep{comanici2025gemini}, GLM-4.5 \citep{zeng2025glm}, DeepSeek-V3 \citep{liu2024deepseek}, Kimi K2-instruct \citep{team2025kimi}, and Qwen3-235b-a22b-Instruct-2507 \citep{yang2025qwen3}. The temperature was set to 0.3, top\_p was set to 0.95, the maximum context length was 128k, and the maximum number of context messages was 200 (When exceeded, the earliest messages will be discarded but the system prompt will be retained). The prompts are available in the Appendix \ref{app.prompt}.

\textbf{Evaluation.} We employ an LLM-as-a-Judge approach to evaluate the results submitted by the agent in a point-wise manner, where each point is assigned a fixed score (refer to the Appendix \ref{app.prompt} for details). The scoring model used is Deepseek-R1.

\textbf{Environments.} As detailed in Section \ref{sec:benchmark}, each environment in \benchname{} exposes a unified action space of various tool types (move, reset, commit, cross, measure, Python coding, notebook, etc.) and multiple turns for investigation on underlying rules.

\subsection{Main Experiments}

Our main experiments evaluate the performance of state-of-the-art LLMs on the \benchname{}, focusing on their ability to handle long-horizon exploration tasks across the three environments: Mystery Grid, Sequence Exploration and Alien Genetics Laboratory. 

We observed that some models submitted their results prematurely, while others submitted too late, which may lead to unfair evaluations when the number of steps is not fixed. Therefore, we consider two settings: one without step restrictions (free) and another with a fixed number of steps (fixed). It is worth noting that the step count referred to here does not indicate the actual number of tool-calling rounds, but rather the number of interactions between the agent and the environment. Basic functionalities such as checking the current state, recording notes, or using the Python interpreter are not counted toward this step limit.

\begin{tcolorbox}[colback=black!5!white, colframe=black!100!black, boxrule=0.5mm, left=0.1mm, right=0.1mm, top=0.1mm, bottom=0.1mm]
\textbf{Takeaway 1.} Some LLMs exhibit a tendency to terminate prematurely rather than sustaining the deliberation and exploration required by long-horizon tasks..
\end{tcolorbox}

\subsubsection{Analysis of Fixed Step Experiments}


\begin{figure*}[htbp]
\centering
\includegraphics[width=\textwidth]{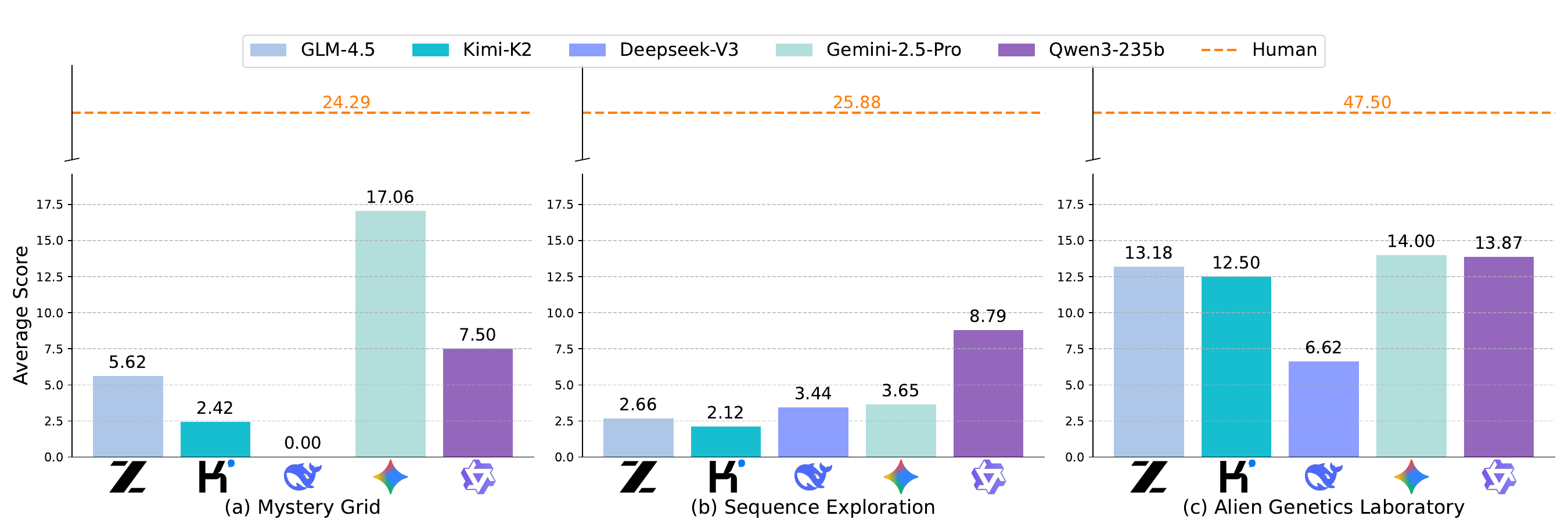}
\vspace{-0.5cm}
\caption{The average scores of various LLMs with fixed steps on \benchname{}.}
\label{fig:main_score}
\end{figure*}

Figure \ref{fig:main_score} presents the average scores of various LLMs with fixed steps (50 for Mystery Grid and Sequence Exploration, 25 for Alien Genetics Laboratory) on \benchname{}, which reveal significant performance gaps across different models and environments. In the Mystery Grid and Alien Genetics Laboratory environments, Gemini-2.5-Pro achieves the highest score, followed by Qwen3-235b and GLM-4.5, while DeepSeek-V3 performs poorly in both scenarios. This suggests that complex tasks involving sustained reasoning and hypothesis testing require agents with stronger reasoning capabilities and better long-term context processing. The second environment, Sequence Exploration, shows the most challenging scenario for all models. The relatively low scores across all models suggest that long-term systematic experimentation and pattern recognition in symbolic domains remain a significant challenge for current LLMs, consistent with observations from \citep{wang2025odysseybench,ling2025longreason} regarding difficulties in long-context reasoning tasks.

\subsubsection{Analysis of Free Step Experiments}

\begin{figure*}[!t]
\centering
\includegraphics[width=\textwidth]{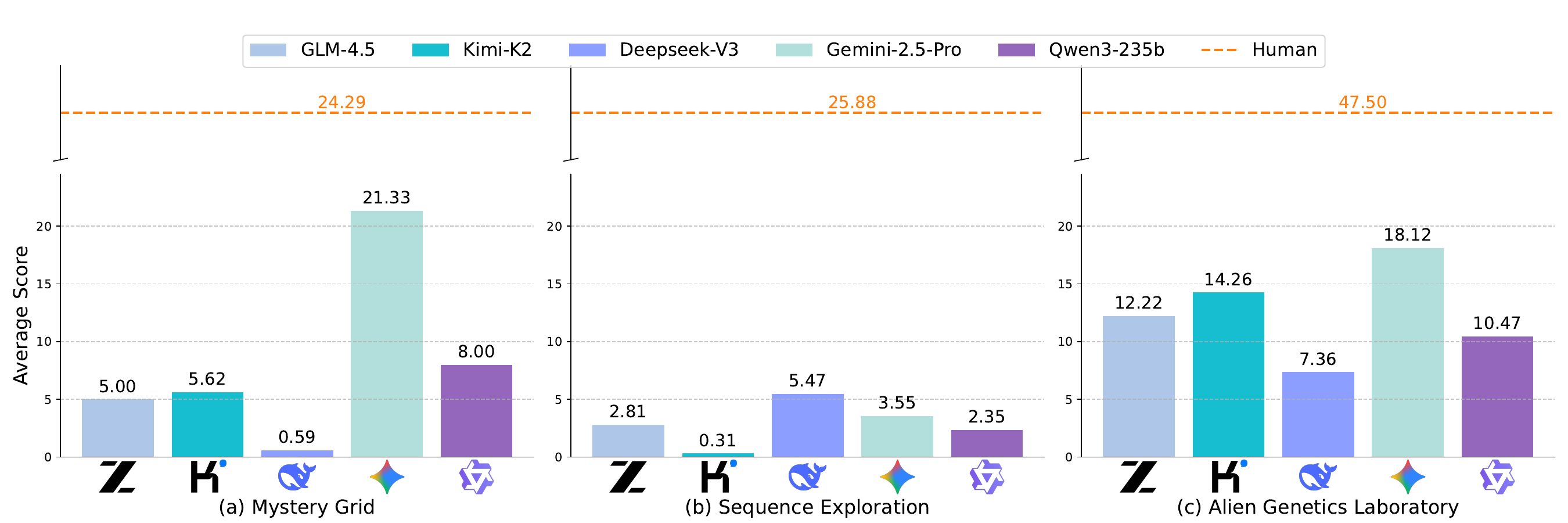}
\vspace{-0.5cm}
\caption{The average scores of various LLMs with free steps on \benchname{}.}
\label{fig:main_free_score}
\end{figure*}

Figure \ref{fig:main_free_score} presents the average scores of various LLMs by removing the constraints on steps, allowing agents to explore until reaching confidence in their hypotheses. 
Removing step constraints resulted in divergent performance patterns across models and environments. Gemini-2.5-Pro demonstrated an improvement of approximately 4 points in Mystery Grid and Alien Genetics Laboratory, suggesting that the initial step constraints were too restrictive for adequate exploration of the environment's rule space. This aligns with findings from \citep{liu2023agentbench} on the value of extended interaction periods. 
In contrast, a decline of 6.44 points for Qwen3-235b in the Sequence Exploration environment suggests that agents may suffer from overconfidence or give up prematurely without further verification. These bidirectional shifts demonstrate that simply providing more exploration opportunities is not universally beneficial. Instead, agents require sophisticated exploration strategies with adequate mechanisms for hypothesis formation, testing, and refinement \citep{zhang2025swe}.

\begin{table}[!t]
\centering
\caption{Comparisons of the average trace length (tokens), tool calls and completion tokens on \benchname{}. The upper and lower panels present the results with fixed steps and free settings. }
\resizebox{\textwidth}{!}{
\begin{tabular}{l ccc ccc ccc}
\toprule
\multirow{2}{*}{Models} & \multicolumn{3}{c}{Mystery Grid} & \multicolumn{3}{c}{Sequence Explore} & \multicolumn{3}{c}{Alien Genetics Laboratory} \\
\cmidrule(lr){2-4} \cmidrule(lr){5-7} \cmidrule(lr){8-10}
&Trace & Toolcalls & Completions &Trace  & Toolcalls & Completions &Trace & Toolcalls & Completions \\
\midrule
\textbf{\textit{Fixed}} \\
GLM-4.5 & 31321 & 87.53 & 17379 & 40339 & 54.00 & 11569 & 31321 & 87.53 & 17378 \\
Kimi-K2 & 29565 & 84.71 & 15627 & 27928 & 66.91 & 8378 & 40328 & 53.65 & 9260 \\
Deepseek-V3 & 23018 & 89.82 & 13650 & 53537 & 92.85 & 30544 & 60847 & 70.75 & 17231 \\
Gemini-2.5-pro & 29357 & 78.56 & 22075 & 64253 & 85.33 & 58125 & 79290 & 69.03 & 29470 \\
Qwen3-235b & 31088 & 89.25 & 18392 & 62015 & 99.30 & 35592 & 52736 & 66.38 & 20054 \\
\midrule
\textbf{\textit{Free}} \\
GLM-4.5  & 32648 & 91.22 & 19611 & 13623 & 17.75 & 7716 & 76434 & 132.54 & 25601\\
Kimi-K2  & 28425 & 63.12 & 14341 & 10507 & 17.87 & 4877 & 21880 & 41.31 & 6728\\
Deepseek-V3  & 30053 & 109.56 & 18952 & 13569 & 21.94 & 7451 & 10582 & 11.69 & 3091\\
Gemini-2.5-pro  & 26215 & 72.18 & 19960 & 16322 & 22.81 & 25262 & 45821 & 32.72 & 32021\\
Qwen3-235b  & 35998 & 138.06 & 11004 & 21480 & 24.76 & 14266 & 17518 & 16.72 & 9290\\
\midrule
\textbf{\textit{Scale (max)}} \\
GLM-4.5 (Naive)  & 96860	&251.57	&55456  &142854 &	181.53	&67574 &209308	&352.20	&79715\\
GLM-4.5 (CRNR)   & 67327 &260.44	&33998  & 120953 &216.50	&60747  &208440	&433.00	&54199 \\
\bottomrule
\end{tabular}}
\label{tab:trace_statistics}
\end{table}

\begin{tcolorbox}[colback=black!5!white, colframe=black!100!black, boxrule=0.5mm, left=0.1mm, right=0.1mm, top=0.1mm, bottom=0.1mm]
\textbf{Takeaway 2.} LLMs markedly underperform humans on \benchname{}, despite often surpassing human baselines on math-reasoning benchmarks.
\end{tcolorbox}

\subsubsection{Analysis on Trace Statistics}
Table \ref{tab:trace_statistics} provides complementary insights into the computational characteristics of agent interactions across both fixed-step and free-step settings.
Fixed-step experiments reveal distinct model strategies under identical constraints. In Sequence Exploration, Qwen3-235b employs relatively more tools (99.30 calls) and long traces (62,015 tokens). In contrast, Gemini-2.5-Pro uses fewer tools (85.33) but generates the most completion tokens (58,125), indicating its higher output efficiency. Similarly, in Alien Genetics Laboratory, Gemini-2.5-Pro produces the longest traces (79,290 tokens), demonstrating superior sustained reasoning within fixed interactions. These differences highlight that various LLMs adopt fundamentally different exploration approaches, e.g., GLM-4.5 and Qwen3-235b may prioritize extensive tool usage, while others like Gemini-2.5-Pro optimize for information generation. 

Free-step experiments expose critical limitations in autonomous exploration calibration, when agents are allowed to determine their own exploration duration. 
In contrast to GLM-4.5, DeepSeek-V3 collapses in Alien Genetics with a shortened trace length (60,847 $\rightarrow$ 10,582 tokens), fewer tool calls (70.75 $\rightarrow$ 11.69), and a marginally improved score in the free-step scenario.
Gemini-2.5-Pro’s adaptive efficiency in free steps suggests potential for improvement. Still, the universal failure in Sequence Exploration highlights a fundamental challenge, where current LLMs may lack intrinsic mechanisms to calibrate the depth of exploration for partially observable, rule-discovery tasks.

\subsection{Ablation Study on Horizon Level}

A natural question is whether the low scores of LLM-based agents stem from the intrinsic difficulty of the tasks in terms of reasoning, or from the long-horizon nature of the environments. To disentangle these factors, we conduct an ablation study focusing on horizon length. Since all three environments in our benchmark share a standard structure, agents must iteratively generate hypotheses, design experiments, observe outcomes, and infer latent rules. We perform the ablation on the \textit{Mystery Grid Environment} as a representative case. This environment offers the most direct and controllable way to vary horizon length by manipulating the number of hidden rules.

Specifically, we adjust the horizon level by varying the number of distinct symbols in the grid from 1 to 5, with each emblem assigned a fixed but randomly chosen effect. Agents are allowed unlimited exploratory attempts, ensuring that performance differences cannot be attributed to step limitations. For fair comparison across different horizon lengths, we normalize scores by dividing the raw reward (20 points per correctly identified rule) by the maximum attainable score ($n \times 20$), and then multiplying by 100.

\begin{table}[h]
\centering
\caption{Ablation results of GLM-4.5 on the Mystery Grid Environment. Horizon level $n$ corresponds to the number of hidden rules. Scores are normalized percentages.}
\label{tab:ablation}
\begin{tabular}{c|ccccc}
\toprule
Horizon level ($n$) & 1 & 2 & 3 & 4 & 5 \\
\midrule
Agent score (\%) & 34.4 & 14.1 & 9.37 & 7.03 & 5.62 \\
Avg. tool calls  & 45.53 & 69.94 & 84.28 & 86.97 & 87.97 \\
\bottomrule
\end{tabular}
\end{table}

Results of GLM-4.5 are reported in Table~\ref{tab:ablation}. As the number of hidden rules increases, performance decreases monotonically, from $34.4$ at $n=1$ to $5.62$ at $n=5$. This sharp decline highlights that agents struggle primarily with sustaining reasoning and memory over extended horizons, rather than with the difficulty of inferring individual rules. This reinforces the conclusion that, rather than the task-intrinsic reasoning difficulty, the horizon length could be the dominant bottleneck for current LLM agents in these environments.

\begin{tcolorbox}[colback=black!5!white, colframe=black!100!black, boxrule=0.5mm, left=0.1mm, right=0.1mm, top=0.1mm, bottom=0.1mm]
\textbf{Takeaway 3.} Agent performance drops sharply as horizon level increases, suggesting that sustaining reasoning over long horizons is a major bottleneck for current LLM agents.
\end{tcolorbox}

\subsection{Analysis of Scaling Experiments}

\subsubsection{Simple Scaling Fails}

To understand how the exploration budget affects agent performance in long-horizon tasks, we conducted scaling experiments by varying the maximum exploration steps within $\{25, 50, 75, 100, 125, 150\}$, except for the Alien Genetics Laboratory which excluded 150 steps. We evaluated GLM-4.5 across all three environments, with each (environment, maximum steps) pair tested on 32 independent runs.

\begin{figure*}[htbp]
\centering
\includegraphics[width=\textwidth]{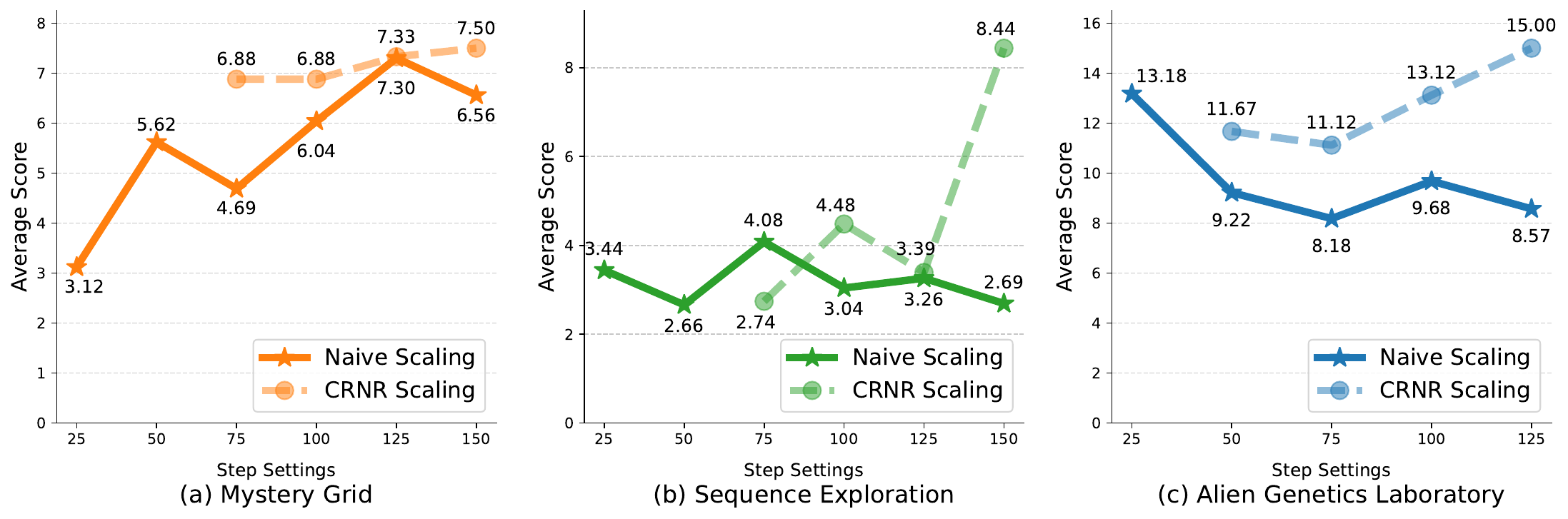}
\vspace{-0.5cm}
\caption{The average scores with scaling steps of GLM 4.5 on \benchname{}.}
\label{fig:scaling_score}
\end{figure*}

As shown in Figure \ref{fig:scaling_score}, the naive scaling approach reveals that simply increasing the interaction steps does not reliably improve performance, with different environments exhibiting distinct optimal step counts before performance declines. In the Mystery Grid environment, scores peak at 125 steps (7.30) before declining to 6.56 at 150 steps, while Alien Genetics Laboratory shows optimal performance at 25 steps before deteriorating. The Sequence Exploration environment demonstrates consistently poor performance across all step budgets, with only marginal improvement to 4.08 at 75 steps. This performance decline after optimal step counts indicates that agents become overwhelmed by excessive information without proper synthesis mechanisms, highlighting that current LLM agents lack the inherent capability to utilize extended interaction budgets for long-horizon tasks effectively.


\subsubsection{A Simple but Effective Scaling Strategy}
To address the limited effectiveness of simply scaling the number of agent–environment interactions, we propose a strategy called \textbf{Context Refresh with Notes Recall} (CRNR). The practical efficiency of CRNR scaling has been verified in Figure \ref{fig:scaling_score}. In this approach, once the accumulated interaction history approaches the model’s context window limit, all prior dialogue turns are cleared except for the system prompt. Then, the agent is instructed to review its self-maintained notes, which serve as an externalized record of essential information from earlier steps. This method operates as a lightweight form of context summarization, enabling the agent to reconstruct necessary knowledge from its notes while avoiding the inefficiencies and confusion caused by excessively long contexts in practice. 

\begin{tcolorbox}[colback=black!5!white, colframe=black!100!black, boxrule=0.5mm, left=0.1mm, right=0.1mm, top=0.1mm, bottom=0.1mm]
\textbf{Takeaway 4.} Simply increasing interaction steps does not reliably improve long-horizon task performance.
\end{tcolorbox}

\section{Why Do Agents Fail in Long-Horizon Task?}
\label{sec:failure}
\subsection{Failure Categorization}

Understanding why agents fail in long-horizon exploration tasks is crucial for identifying their fundamental limitations and guiding future improvements.
Based on manual trajectory analysis, we propose a two-level framework to systematically categorize failures according to their \textit{origin} (root causes) and \textit{manifestation} (observable patterns). 


\subsection{Failure Origin: Root Causes of Agent Breakdown}

\begin{figure*}[!t]
\centering
\includegraphics[width=\textwidth]{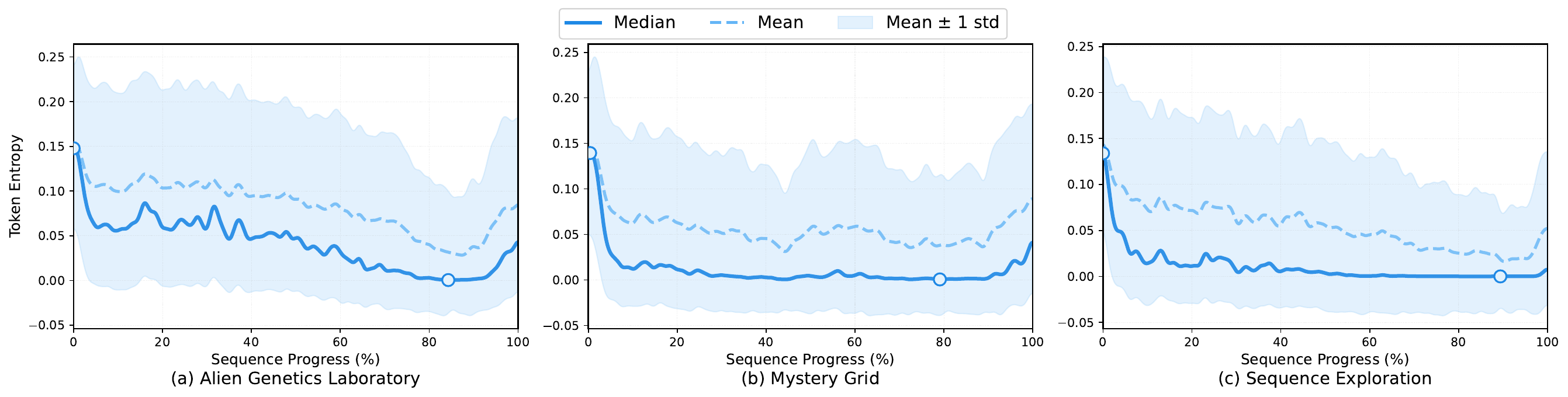}
\vspace{-0.5cm}
\caption{The token entropy dynamics of GLM-4.5 on \benchname{}.}
\label{fig:entropy}
\end{figure*}

\paragraph{In-context Locking.} 
This is a \textit{process-induced root cause}, where agents become locked on initial patterns, assumptions, or habits, and lack mechanisms for dynamic adjustment, exploration, or self-reflection.
As a result, they persist with suboptimal strategies, or explore only within a narrow band of familiar actions without genuine adaptation throughout the trajectory.
We also conducted a statistical analysis of this phenomenon. Specifically, we plotted the entropy dynamics of GLM-4.5 across three distinct environments as a function of sequence position in Figure \ref{fig:entropy}. The median entropy, which was chosen over the mean to reduce sensitivity to extreme values, exhibits a consistent downward trend, suggesting the emergence of in-context locking. Toward the end of the sequence, entropy increases. This rise can be attributed to the model producing its final output at the conclusion of the interaction, a stage that is often associated with heightened uncertainty.

\paragraph{Foundational Capability Gaps.} 
In contrast, this is a \textit{capacity-induced root cause}, rooted in the intrinsic limitations of the underlying model. Even without early-context inertia, the agent may falter because it lacks essential competencies such as robust logical reasoning, reliable long-term memory management, stable tool calling, or effective multi-step planning.

\subsection{Failure Manifestation: How Failures Surface in Trajectories}

While the origins explain why agents fail, their concrete manifestations determine how such failures unfold in trajectories. 
We categorize observable patterns in Table \ref{tab:failure_manifestation_flat}, with detailed definitions and specific case studies presented in Appendix \ref{app:failure}. By distinguishing failure origins from their observable manifestations, this framework provides a lens for a clear diagnosis and attribution.

\begin{tcolorbox}[colback=black!5!white, colframe=black!100!black, boxrule=0.5mm, left=0.1mm, right=0.1mm, top=0.1mm, bottom=0.1mm]
\textbf{Takeaway 5.} Agents tend to get stuck in early assumptions and narrow strategies, with entropy analysis confirming this \textbf{in-context locking} effect.
\end{tcolorbox}

\begin{wrapfigure}{r}{0.4\textwidth}
\centering
\includegraphics[width=\linewidth]{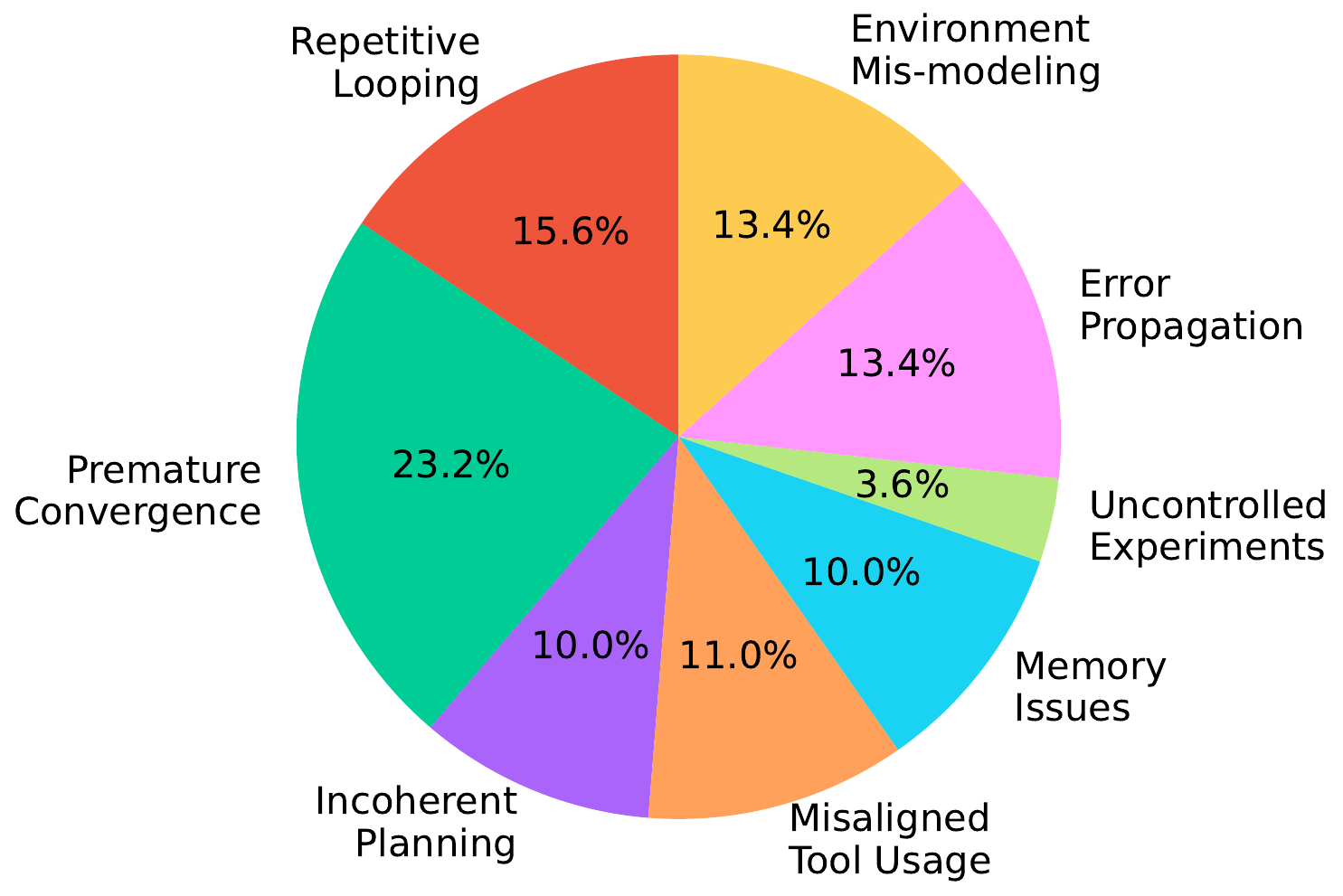}
\vspace{-0.5cm}
\caption{The categories and distribution of agents' failures on \benchname.}
\label{fig:right}
\end{wrapfigure}


\section{Conclusion}
\label{sec:conclusion}

This work introduces \benchname, a benchmark specifically designed to evaluate language agents in long-horizon, partially observable environments. Unlike prior efforts that emphasize short trajectories or fully observable tasks, \benchname\ captures real-world complexity by requiring sustained reasoning, planning, memory management, and tool use across extended interactions, reaching on average \textbf{200k+} tokens and \textbf{400+} tool calls under the heaviest setting (while standard runs still exceed \textbf{35k} tokens and \textbf{60}+ tool calls). Our experiments show that state-of-the-art agents exhibit substantial limitations, whereas human participants attain markedly higher success rates, underscoring a persistent capability gap. Trajectory-level analyses further reveal recurring error patterns rooted in in-context locking and foundational skill deficiencies. Taken together, these findings indicate that progress in long-horizon exploration will require advances beyond simple scaling, pointing toward principled memory integration, adaptive reasoning, and robust exploration strategies.

\clearpage
\newpage

\section*{Ethics Statements}
\label{sec:a0}


This research presents \benchname{}, a benchmark designed to evaluate agent capabilities in long-horizon, partially observable scenarios. The benchmark environments are synthetically constructed and do not involve sensitive data or real-world identities. All experiments are conducted in simulated settings with explicitly defined rules, ensuring that no ethical risks are associated with privacy, bias, or harmful content. 
Partial evaluation was conducted with the assistance of human participants. Prior to participation, all human participants were provided with detailed information about the research purpose, procedures, and data usage.
The research was conducted in accordance with the principles outlined in the Declaration of Helsinki and underwent an informal ethics review process performed by the authors' peers.
No personally identifiable information was collected during the evaluations, and all data was anonymized to ensure participant anonymity. Participants were informed of their right to withdraw from the study at any time without penalty.

All sources are properly cited in accordance with academic standards. The authors confirm that this work was conducted in accordance with the principles of academic integrity and research ethics.

\section*{Reproducibility Statement}

The prompts for three environments (Mystery Grid, Sequence Exploration, and Alien Genetics Laboratory), agent code, and evaluation scripts are available in the attached supplementary files. Experiment configurations, including hyperparameters, random seeds, and model versions (Gemini-2.5-Pro, GLM-4.5, DeepSeek-V3, Kimi K2-instruct, Qwen3-235b) are fully documented. All results were generated using fixed computational resources detailed in Section 5 and Appendices A-D. This enables independent verification of all findings.

\bibliography{iclr2026_conference}
\bibliographystyle{iclr2026_conference}

\clearpage
\newpage

\appendix

\begin{center}
\Large 
Appendix 
\end{center}

\startcontents[sections]  
\printcontents[sections]{}{1}{\setcounter{tocdepth}{2}}  
\vskip 0.2in
\hrule

\section{Environmental Settings}
\label{sec:a1}
\label{app.html}

The detailed settings of the three environments in \benchname{} with graphical user interface (GUI) visualizations are illustrated below.

\begin{figure*}[htbp]
\centering
\includegraphics[width=0.95\textwidth]{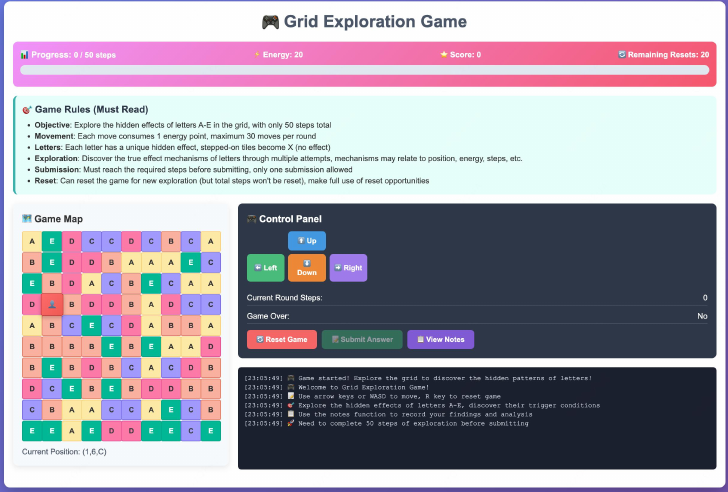}
\caption{The graphical user interface of the Mystery Grid environment.}
\label{fig:gridhtml}
\end{figure*}

\begin{table*}[htbp]
\centering 
\caption{Environmental settings in the Mystery Grid environment.}
\resizebox{\textwidth}{!}{
\begin{tabular}{|p{0.25\textwidth}|p{0.75\textwidth}|}
\hline
\multicolumn{2}{|c|}{\textbf{Introduction}} \\
\hline
\multicolumn{2}{|p{\textwidth}|}{
The LLM agent navigates a 10×10 grid to discover hidden rules governing five letters (A-E). Operating with 20 initial energy points and consuming 1 per move, the agent must complete a required number of steps (typically 50) before submitting conclusions. The core challenge involves identifying consistent, non-observable mechanistic patterns through systematic exploration, rather than maximizing score.
}\\
\hline
\multicolumn{2}{|c|}{\textbf{Rules}} \\
\hline
\multicolumn{2}{|p{\textwidth}|}{1) \textbf{\textit{Letter A:}} Step count effect. When the total step count is divisible by 3, the agent gains 2 points; otherwise, it loses 1 point.}\\
\multicolumn{2}{|p{\textwidth}|}{2) \textbf{\textit{Letter B:}} Corner and edge effects. When positioned at corners (coordinates where x and y are both 0 or 9) or edges (coordinates where x or y is 0, 1, 8, or 9), the agent gains 3 points.}\\
\multicolumn{2}{|p{\textwidth}|}{3) \textbf{\textit{Letter C:}} Visit count-based effect. When the cumulative visit count for that specific letter type is odd, the agent gains 1 point; when even, no points are awarded. }\\
\multicolumn{2}{|p{\textwidth}|}{4) \textbf{\textit{Letter D:}} Energy management effect. When energy is below 10 points, the agent loses 2 points but gains 10 energy; when energy is 10 or above, the agent gains 1 point. }\\
\multicolumn{2}{|p{\textwidth}|}{5) \textbf{\textit{Letter E:}} Complex position calculation effect. The agent's score changes by the value of ($x-y$), where x represents the column coordinate and y represents the row coordinate. }\\
\hline
\multicolumn{1}{|c|}{\textbf{Given Tools}} &  \multicolumn{1}{|c|}{\textbf{Tool descriptions}} \\
\hline
1) Move & \multicolumn{1}{|p{0.70\textwidth}|}{1) Move in one of four cardinal directions (up, down, left, right) on the grid and update the state.} \\
\hline
2) Get Current State & \multicolumn{1}{|p{0.70\textwidth}|}{2) Provides the agent's current position $(x,y)$, energy level, score, step count, and information about nearby tiles within a two-tile radius.} \\
\hline
3) Get Full Map & \multicolumn{1}{|p{0.70\textwidth}|}{3) Provides the complete grid state with coordinates and letter positions, allowing the agent to plan exploration routes strategically.} \\
\hline
4) Reset & \multicolumn{1}{|p{0.70\textwidth}|}{4) Enables the agent to restart the game with a new random grid and start location. This tool is limited to a maximum of 20 uses and is essential for validating hypotheses through investigation.}  \\
\hline
5) Python Interpreter & \multicolumn{1}{|p{0.70\textwidth}|}{5) Allows the agent to utilize Python coding to solve some computation problems.} \\
\hline
6) Write Note & \multicolumn{1}{|p{0.70\textwidth}|}{6) Append a concise note (e.g., hypotheses, observations, partial results) to a persistent scratchpad for the current episode.} \\
\hline
7) Check Note & \multicolumn{1}{|p{0.70\textwidth}|}{7) Retrieve previously written notes, to verify hypotheses, avoid repeated mistakes, and support controlled experiments.} \\
\hline
8) Commit Final Result & \multicolumn{1}{|p{0.70\textwidth}|}{8) Allows the agent to submit its final analysis of the letter effects. This tool can only be used once, after the required total steps have been completed.} \\
\hline
\end{tabular}
}
\end{table*}

\begin{figure*}[htbp]
\centering
\includegraphics[width=0.95\textwidth]{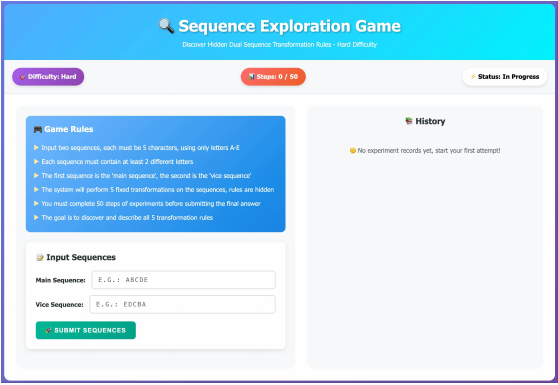}
\caption{The graphical user interface of the Sequence Explore environment.}
\label{fig:seqhtml}
\end{figure*}

\begin{table*}[htbp]
\centering 
\caption{Environmental settings in the Sequence Explore environment.}
\resizebox{\textwidth}{!}{
\begin{tabular}{|p{0.25\textwidth}|p{0.75\textwidth}|}
\hline
\multicolumn{2}{|c|}{\textbf{Introduction}} \\
\hline
\multicolumn{2}{|p{\textwidth}|}{
A sequence exploration game where the agent uncovers five hidden transformation rules by inputting pairs of letter sequences and observing deterministic and periodic changes, then submits the deduced mechanisms through controlled experimentation.
}\\
\hline
\multicolumn{2}{|c|}{\textbf{Rules}} \\
\hline
\multicolumn{2}{|p{\textwidth}|}{1) \textbf{\textit{Rule 1:}} Interleaving or merging characters from main and vice sequences, with the leading sequence possibly changing based on step count or other factors.}\\
\multicolumn{2}{|p{\textwidth}|}{2) \textbf{\textit{Rule 2:}} Character position-based operations, potentially involving arithmetic on positions (e.g., addition, modular arithmetic) or alphabetic shifts.}\\
\multicolumn{2}{|p{\textwidth}|}{3) \textbf{\textit{Rule 3:}} Position-dependent selection or copying (e.g., step-number modulo patterns) where specific characters are duplicated or extracted.}\\
\multicolumn{2}{|p{\textwidth}|}{4) \textbf{\textit{Rule 4:}} Character-wise combination between sequences, possibly involving comparisons (max/min) or operations using historical states (e.g., previous main sequences).}\\
\multicolumn{2}{|p{\textwidth}|}{5) \textbf{\textit{Rule 5:}} Frequency-based transformations where occurrence patterns drive modifications (e.g., replacing the most frequent character under prime-numbered steps).}\\
\hline
\multicolumn{1}{|c|}{\textbf{Given Tools}} &  \multicolumn{1}{|c|}{\textbf{Tool descriptions}} \\
\hline
1) Input Sequences & \multicolumn{1}{|p{0.70\textwidth}|}{1) Submit pairs of fixed-length letter sequences for transformation and receive detailed results of all applied rules at the current step.} \\
\hline
2) Python Interpreter & \multicolumn{1}{|p{0.70\textwidth}|}{2) Allows the agent to utilize Python coding to solve computation or verification tasks.} \\
\hline
3) Write Note & \multicolumn{1}{|p{0.70\textwidth}|}{3) Append a concise note (e.g., hypotheses on rules, observations, partial results) to a persistent scratchpad for the current episode.} \\
\hline
4) Check Note & \multicolumn{1}{|p{0.70\textwidth}|}{4) Retrieve previously written notes, optionally filtered by keywords or indices, to verify hypotheses, avoid repeated mistakes, and support controlled experiments.} \\
\hline
5) Commit Final Result & \multicolumn{1}{|p{0.70\textwidth}|}{5) Submit the inferred mechanisms of the five rules for evaluation; allowed only once after reaching the required steps.} \\
\hline
\end{tabular}
}
\end{table*}

\begin{figure*}[htbp]
\centering
\includegraphics[width=0.95\textwidth]{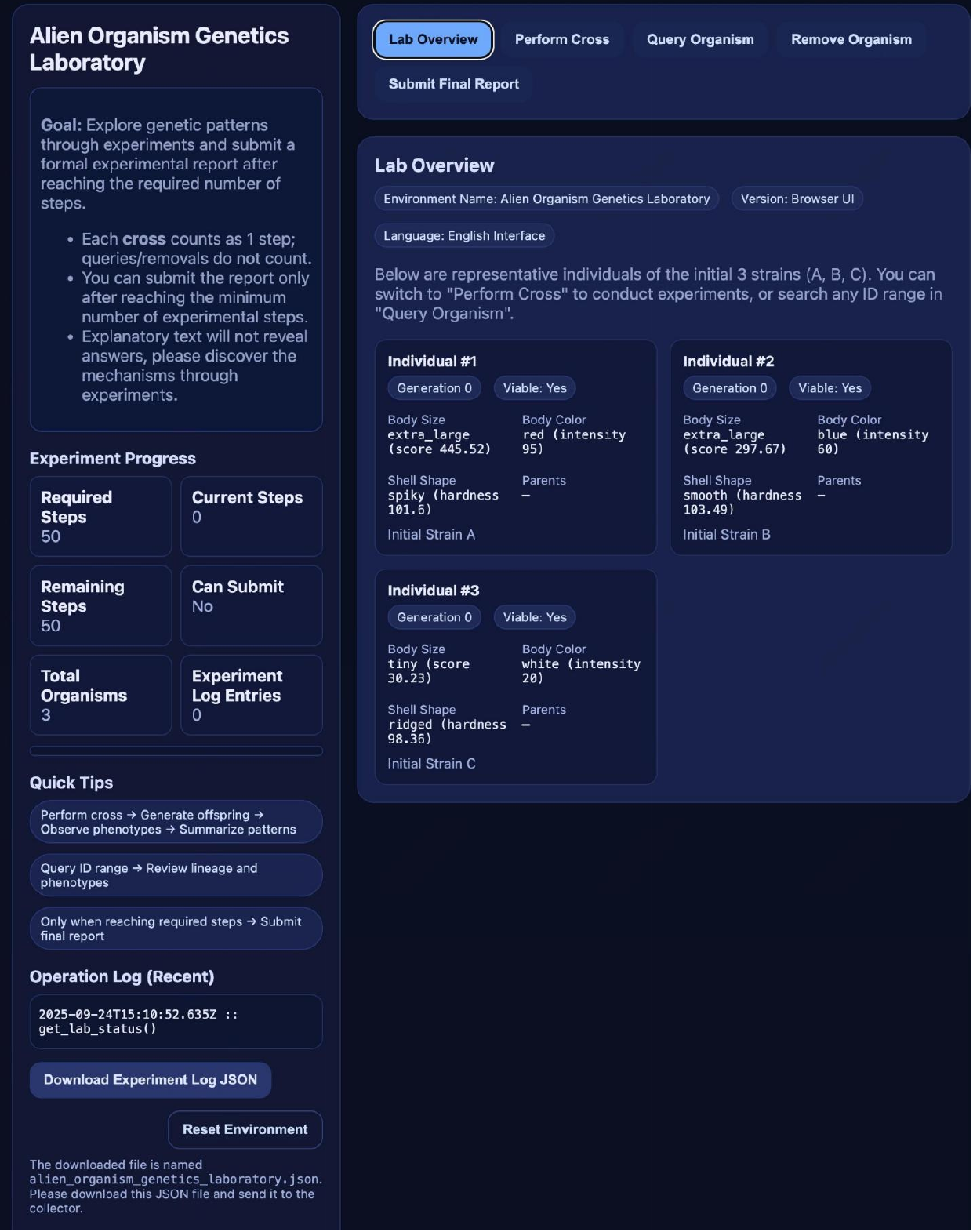}
\caption{The graphical user interface of Alien Genetics Laboratory environment.}
\label{fig:biohtml}
\end{figure*}

\begin{table*}[!t]
\centering 
\caption{Environmental settings in the Alien Genetics Laboratory environment.}
\resizebox{\textwidth}{!}{
\begin{tabular}{|p{0.25\textwidth}|p{0.75\textwidth}|}
\hline
\multicolumn{2}{|c|}{\textbf{Introduction}} \\
\hline
\multicolumn{2}{|p{\textwidth}|}{
An Alien Genetics Laboratory experiment where the agent discovers inheritance rules of triploid alien organisms through controlled crosses, analyzes trait patterns, and identifies lethal gene combinations, before submitting final conclusions.
}\\
\hline
\multicolumn{2}{|c|}{\textbf{Rules}} \\
\hline
\multicolumn{2}{|p{\textwidth}|}{1) \textbf{\textit{Ploidy Level:}} Organisms are triploid; each gene locus carries three alleles rather than the two typical of diploids.}\\
\multicolumn{2}{|p{\textwidth}|}{2) \textbf{\textit{Meiotic Segregation Mechanism:}} Special meiosis with unequal 1\,+\,2 segregation produces haploid and diploid gametes.}\\
\multicolumn{2}{|p{\textwidth}|}{3) \textbf{\textit{Viability Constraint:}} Only triploid zygotes are viable; other ploidy outcomes are lethal.}\\
\multicolumn{2}{|p{\textwidth}|}{4) \textbf{\textit{Body Size Inheritance Pattern:}} Body size follows an additive dosage effect; alleles contribute quantitatively to phenotype.}\\
\multicolumn{2}{|p{\textwidth}|}{5) \textbf{\textit{Body Size Allele Identification:}} Three distinct alleles (S1, S2, S3) control size with different quantitative effects.}\\
\multicolumn{2}{|p{\textwidth}|}{6) \textbf{\textit{Body Size Quantitative Values:}} Approximate allele values: 200$\pm$20, 50$\pm$10, 10$\pm$5.}\\
\multicolumn{2}{|p{\textwidth}|}{7) \textbf{\textit{Color Dominance Hierarchy:}} Color follows strict dominance C1 (red) $>$ C2 (blue) $>$ C3 (white).}\\
\multicolumn{2}{|p{\textwidth}|}{8) \textbf{\textit{Color Complete Dominance:}} Phenotype is determined solely by the most dominant allele present.}\\
\multicolumn{2}{|p{\textwidth}|}{9) \textbf{\textit{Shell Shape Cyclic Dominance:}} Shell alleles show cyclic dominance: H1 (spiky) $>$ H2 (smooth) $>$ H3 (ridged) $>$ H1.}\\
\multicolumn{2}{|p{\textwidth}|}{10) \textbf{\textit{Shell Lethal Combination:}} Genotype H1\,+\,H2\,+\,H3 is lethal and non-viable.}\\
\hline
\multicolumn{1}{|c|}{\textbf{Given Tools}} &  \multicolumn{1}{|c|}{\textbf{Tool descriptions}} \\
\hline
1) Conduct Cross & \multicolumn{1}{|p{0.70\textwidth}|}{1) Perform a controlled cross between two organisms; return offspring phenotypes, viability rates, and lethality statistics.} \\
\hline
2) Query Organisms & \multicolumn{1}{|p{0.70\textwidth}|}{2) Examine organisms within a specified ID range (e.g., \texttt{start\_id} to \texttt{end\_id}) for traits and lineage; supports optional phenotype/genotype filters.} \\
\hline
3) Get Lab Status & \multicolumn{1}{|p{0.70\textwidth}|}{3) Provide current laboratory resource usage, experiment progress, and remaining experiment budget.} \\
\hline
4) Remove Organisms & \multicolumn{1}{|p{0.70\textwidth}|}{4) Remove specified organisms to manage capacity constraints.} \\
\hline
5) Python Interpreter & \multicolumn{1}{|p{0.70\textwidth}|}{5) Allow the agent to utilize Python coding for calculations, simulations, or verification.} \\
\hline
6) Write Note & \multicolumn{1}{|p{0.70\textwidth}|}{6) Append a concise note (e.g., hypotheses, observations, partial results) to a persistent scratchpad for the current episode.} \\
\hline
7) Check Note & \multicolumn{1}{|p{0.70\textwidth}|}{7) Retrieve previously written notes, optionally filtered by keywords or IDs, to verify hypotheses, avoid repeated mistakes, and support controlled experiments.} \\
\hline
8) Commit Final Result & \multicolumn{1}{|p{0.70\textwidth}|}{8) Submit the final conclusions describing inferred inheritance rules; allowed only once after reaching the required steps.} \\
\hline
\end{tabular}
}
\end{table*}


\clearpage
\newpage

\section{Prompt and Annotation Details}
\label{sec:a2}
\label{app.prompt}

\subsection{Agent Prompt for Mystery Grid Environment}

\begin{tcolorbox}[enhanced,breakable,colback=blue!5,colframe=blue!80,fonttitle=\bfseries]
GAME OBJECTIVE:

You are an agent in a 10x10 grid world. Your goal is to discover the hidden rules of effects of different letters (A-E).\\

BASIC RULES:

- You start at initial position with 20 energy points

- Each move costs 1 energy point

- You have a maximum of 30 steps per game round, and a required total steps of {required\_steps}, you cannot move after reaching the limit, and you cannot commit before reaching the limit.

- The grid contains letters A through E, each with unique hidden effects on your score or energy. X means you have already stepped on this tile and it has no effect.

- The effects of each letter is consistent across games, but you need to discover them through exploration.

- After you step on a tile, its letter changes to 'X'.

- The game ends when you run out of energy or reach 30 steps.\\

IMPORTANT NOTES:

- Letter effects are HIDDEN but follow CONSISTENT rules across different games and you must discover the rules through experimentation. The effects may be related to your position, energy, step count or visit count. 

- Each letter (A-E) has a different scoring mechanics that you need to figure out. The effect of each tile should not be concluded from a single try; multiple validations under different situations are essential to confirm that the observed outcome is not merely a random occurrence. You should fully utilize the reset opportunity to explore the effects of each letter.

- Use notes to record your plans, ideas, and discoveries, and review them frequently to stay on track. Also, record the effects into notes of each letter as you discover them. You may also need to record the changes after each move for later analysis.

- You can reset and replay multiple times to learn the patterns. Your goal is not to maximize the score, but to explore the mechanism clearly. It is recommended to make full use of each game until all steps or energy are exhausted, avoiding resets too easily. At the same time, repeated trials should be conducted to ensure the conclusions are correct before committing. You can only commit your answer once and then the game ends, commit only when you are confident about the effects of each letter.

- Since each step consumes 1 energy, it is therefore not counted as an effect.

COMMIT REQUIREMENTS:

- Complete coverage: You must provide exactly one effect description for each of A, B, C, D, and E.

- Trigger condition required: Your description must specify the underlying trigger condition, which may depend on: step count, visit count, energy, grid position. Stating only surface patterns such as “sometimes +2, sometimes -1” is not sufficient.

- No vague descriptions: Do not use terms like “random,” “uncertain,” or “complex.” Do not provide multiple conflicting explanations.

- Evaluation is strict: only if you clearly identify the correct trigger condition and effect will your answer be marked as correct.

COORDINATE FORMAT:

All positions are shown as (x,y,letter) where x=column, y=row, letter=current tile letter

Now start exploring and discovering the hidden mechanics.

\end{tcolorbox}

\subsection{Agent Prompt for Sequence Explore Environment}

\begin{tcolorbox}[enhanced,breakable,colback=blue!5,colframe=blue!80,fonttitle=\bfseries]
GAME OBJECTIVE:

You are an agent facing a hidden dual-sequence transformation system. Your goal is to discover the hidden rules applied to pairs of sequences of letters (A-E).

BASIC RULES:\\
- You must input TWO sequences of EXACTLY 5 CHARACTERS each, using only letters A-E\\
- Each sequence must contain at least 2 different letters from A-E\\
- The first sequence is the "main str" and the second is the "vice str"\\
- Each input pair consumes 1 step. There is a required total steps of $\{required\_steps\}$ to complete the game\\
- After inputting a sequence pair, they will undergo a fixed chain of transformations\\
- All transformations apply the same rules consistently, operating on current results and/or original inputs\\
- The same hidden rules apply consistently across the game, but you must discover them through exploration\\
- Game ends when you reach the maximum step count or submit your final answer\\

IMPORTANT NOTES:
- Effects are deterministic but hidden; repeated testing is needed to confirm true rules\\
- Each rule may depend on sequence content, position, history, or relationships between main/vice\\
- Use notes to record observed transformations and hypotheses\\
- Do not commit your final answer until you are confident, as you can only commit once\\
- The goal is to uncover and describe the mechanisms behind dual-sequence transformations\\

SEQUENCE FORMAT:\\
Inputs are pairs like ("ABCDE", "DECBA"), ("AABCE", "DDEAC"). Each must be exactly 5 characters with at least 2 different letters.\\
Outputs are shown step by step with rule applications.\\

DIFFICULTY LEVELS:\\
- Easy: Uses simplified versions of rules 1-5\\
- Hard: Uses complex versions of rules 1-5\\

Now begin experimenting and uncover the transformation rules.
\end{tcolorbox}

\subsection{Agent Prompt for Genetics Lab Environment}

\begin{tcolorbox}[enhanced,breakable,colback=blue!5,colframe=blue!80,fonttitle=\bfseries]
GAME OBJECTIVE:

You are an agent in a 10x10 grid world. Your goal is to discover the hidden rules of effects of different letters ($\{letter\_range\}$).

BASIC RULES: \\
- You start at initial position with 20 energy points \\
- Each move costs 1 energy point\\
- You have a maximum of 30 steps per game round{required\_steps\_text}\\
- The grid contains letters $\{letter\_range\}$, each with unique hidden effects on your score or energy. X means you have already stepped on this tile and it has no effect.\\
- The effects of each letter is consistent across games, but you need to discover them through exploration.
- After you step on a tile, its letter changes to 'X'\\
- The game ends when you run out of energy or reach 30 steps\\

IMPORTANT NOTES:\\
- Letter effects are HIDDEN but follow CONSISTENT rules across different games and you must discover the rules through experimentation. The effects may be related to your position, energy, step count or visit count. \\
- Each letter ($\{letter\_range\}$) has a different scoring mechanics that you need to figure out. The effect of each tile should not be concluded from a single try; multiple validations under different situations are essential to confirm that the observed outcome is not merely a random occurrence. You should fully utilize the reset opportunity to explore the effects of each letter.\\
- Use notes to record your plans, ideas, and discoveries, and review them frequently to stay on track. Also, record the effects into notes of each letter as you discover them. You may also need to record the changes after each move for later analysis.\\
- You can reset and replay multiple times to learn the patterns. Your goal is not to maximize the score, but to explore the mechanism clearly. It is recommended to make full use of each game until all steps or energy are exhausted, avoiding resets too easily. At the same time, repeated trials should be conducted to ensure the conclusions are correct before committing. You can only commit your answer once and then the game ends, commit only when you are confident about the effects of each letter.\\
- Since each step consumes 1 energy, it is therefore not counted as an effect.\\

COMMIT REQUIREMENTS:\\
- Complete coverage: You must provide exactly one effect description for each of $\{letter\_range\}$.\\
- Trigger condition required: Your description must specify the underlying trigger condition, which may depend on: step count, visit count, energy, grid position. Stating only surface patterns such as "sometimes +2, sometimes -1" is not sufficient.\\
- No vague descriptions: Do not use terms like "random," "uncertain," or "complex." Do not provide multiple conflicting explanations.\\
- Evaluation is strict: only if you clearly identify the correct trigger condition and effect will your answer be marked as correct.\\

COORDINATE FORMAT:\\
All positions are shown as (x,y,letter) where x=column, y=row, letter=current tile letter\\

Now start exploring and discovering the hidden mechanics.\\
\end{tcolorbox}

\subsection{LLM Judge Prompt for Mystery Grid Environment}
\begin{tcolorbox}[enhanced,breakable,colback=blue!5,colframe=blue!80,fonttitle=\bfseries]

You are an evaluator. Compare the agent's submitted mapping of letters to effect descriptions against the ground truth mapping.\\
The agent should have inferred what each letter (A, B, C, D, E) does when stepped on.\\

$<$Ground truth mapping$>$:\\
\{ground\_truth\}

$<$Agent submitted mapping$>$:\\
\{content\}

Compare the agent's descriptions with the ground truth effects. A mapping is considered correct if the agent's description captures the essential rule of the effect, even if the wording is different. Since each step consumes 1 energy, it is therefore not counted as an effect. However, it is also correct if the agent specifies a consumption of 1 energy for each effect. You should follow these rules to judge:\\
Evaluate correctness according to these strict rules:\\

1. **Trigger Condition Must Be Identified**\\
An effect description is correct **only if** the agent identifies the underlying *trigger condition* (e.g., step count, visit count parity, energy level, or coordinate-based calculation).\\
Simply describing observed outcome patterns (e.g., “sometimes +2, sometimes -1”, “alternating pattern”, “mostly +1 with exceptions”) **without stating the actual trigger condition** is incorrect.\\

2. **Equivalent Wording Allowed**\\
Different wording is acceptable only if it expresses the *same logical condition*.\\
Example: “odd/even visits” $=$ “visit count parity”.\\
Example: “boundary positions” $=$ “corners and edges”.\\
Descriptions that skip the explicit condition and only give surface observations are incorrect.\\

3. **Vague or Wrong Attributions are Incorrect**\\
If the agent uses vague terms such as “variable”, “random”, “uncertain”, or “complex behavior” instead of specifying the true condition, it is incorrect.\\
If the agent attributes the effect to the wrong factor (e.g., says “position-dependent” when the actual rule depends on step count or energy), it is incorrect.\\

4. **One-to-One Mapping Requirement**\\
Each symbol must map to exactly one effect rule.\\
Multiple conflicting explanations, incomplete conditions, or missing mappings are incorrect.\\

5. **Evaluation Criteria**\\
Mark **Correct** only if:\\
The trigger condition identified by the agent matches the ground truth condition.\\
The outcome description aligns with the ground truth rule.\\
Otherwise, mark **Incorrect**.\\

Format your answer as json:

\begin{jsoncode}
    {
  "final_score": 40,
  "score_breakdown": [
    {"criterion": "A", "max_score": 20, "awarded_score": 20, "comment": <your comment>},
    {"criterion": "B", "max_score": 20, "awarded_score": 0, "comment": <your comment>},
    {"criterion": "C", "max_score": 20, "awarded_score": 0, "comment": <your comment>},
    {"criterion": "D", "max_score": 20, "awarded_score": 20, "comment": <your comment>},
    {"criterion": "E", "max_score": 20, "awarded_score": 0, "comment": <your comment>},
  ]
}
\end{jsoncode}

\end{tcolorbox}

\subsection{LLM Judge Prompt for Sequence Explore Environment}
\begin{tcolorbox}[enhanced,breakable,colback=blue!5,colframe=blue!80,fonttitle=\bfseries]
You are an evaluator. Compare the agent's submitted description of transformation rules against the ground truth rules.\\
The agent should have inferred what each rule does in the sequence transformation chain.\\

$<$Ground truth rules$>$:
\{ground\_truth\}

$<$Agent submitted description$>$:
\{content\}

Compare the agent's descriptions with the ground truth rules. A rule description is considered correct if the agent's explanation captures the essential mechanism of the rule, even if the wording is different. Each rule has 20 points, for a total of 100 points. Provide a detailed breakdown of scores for each rule and the final score.\\

Evaluate correctness according to these strict rules:

1. **Mechanism Must Be Identified**\\
A rule description is correct **only if** the agent identifies the underlying *transformation mechanism* (e.g., character interleaving, position-based operations, step-count dependencies, history references).\\
Simply describing observed patterns without stating the actual mechanism is incorrect.\\

2. **Equivalent Wording Allowed** \\
Different wording is acceptable only if it expresses the *same logical transformation*.\\
Example: "alternating characters" = "interleaving main and vice".\\
Example: "reverse and shift" = "flip sequence and advance letters".\\
Example: "current step number" = "total steps" / "num of total experiments"\\

3. **Vague or Wrong Mechanisms are Incorrect**\\
If the agent uses vague terms such as "complex pattern", "variable behavior", or "depends on context" without specifying the true mechanism, it is incorrect.\\
If the agent attributes the rule to the wrong mechanism, it is incorrect (0 score should be assigned).\\

4. **Complete Rule Chain Required**\\
Each rule in the sequence must be correctly identified and described.\\
Missing rules, incomplete mechanisms, or conflicting explanations are incorrect.\\

5. **Evaluation Criteria**\\
Mark **Correct** only if:\\
- The transformation mechanism identified matches the ground truth rule.\\
- The description explains how inputs are transformed to outputs.\\
- The agent shows understanding of when/how the rule applies.\\

Format your answer as json:\\
\begin{jsoncode}
    {
  "final_score": 30,
  "score_breakdown": [
    {"criterion": "rule_1", "max_score": 20, "awarded_score": 10, "comment": <your comment>},
    {"criterion": "rule_2", "max_score": 20, "awarded_score": 0, "comment": <your comment>},
    {"criterion": "rule_3", "max_score": 20, "awarded_score": 0, "comment": <your comment>},
    {"criterion": "rule_4", "max_score": 20, "awarded_score": 20, "comment": <your comment>},
    {"criterion": "rule_5", "max_score": 20, "awarded_score": 0, "comment": <your comment>},
  ]
}
\end{jsoncode}

\end{tcolorbox}

\subsection{LLM Judge Prompt for Genetics Lab Environment}
\begin{tcolorbox}[enhanced,breakable,colback=blue!5,colframe=blue!80,fonttitle=\bfseries]
You are an evaluator. Compare the agent's submitted description of inheritance rules against the scoring rules.

Scoring rules:\\
\#\# A. Fundamental Understanding (25 points)\\
1. Recognition of triploidy - Identifies that the organism is triploid (three alleles per locus). *(Medium, 15 points)* \\
2. Meiosis process - Correctly describes the special meiosis mechanism (gametes are 1n and 2n due to unequal segregation). *(Easy, 5 points)*
3. Viability constraint - Explains that only triploid zygotes are viable; other ploidy combinations are lethal. *(Easy, 5 points)* \\

\#\# B. Body Size Trait (Dosage Effect) (35 points)\\
4. Inheritance rule - Discovers that body size follows an additive dosage effect. *(Easy, 5 points)*
5. Allele identification - Correctly identifies the existence of three distinct alleles for size. *(Medium, 10 points)*\\
6. Quantitative description - Provides accurate quantitative values for the three alleles (approx. 200 ± 20; 50 ± 10; 10 ± 5). *(Hard, 20 points)*\\

\#\# C. Color Trait (Dominance Hierarchy) (10 points)\\
7. Dominance hierarchy - Identifies that color follows a strict dominance order: Red (C1) $>$ Blue (C2) $>$ White (C3). *(Easy, 5 points)*\\
8. Complete dominance - States that the dominance is complete: phenotype intensity is determined solely by the dominant allele. *(Easy, 5 points)*\\

\#\# D. Shell Trait (Cyclic Interaction \& Lethality) (30 points)\\
9. Cyclic dominance - Identifies that shell shape alleles follow a cyclic dominance relationship: Spiky (H1) $>$ Smooth (H2) $>$ Ridged (H3) $>$ Spiky. *(Medium, 10 points)*\\
10. Lethal combination - Explains that the presence of all three shell alleles (H1 + H2 + H3) is lethal. *(Hard, 20 points)*\\

$<$Agent submitted report$>$:
\{content\}

Evaluation guidelines:

1. **Mechanism Identification Required**\\
For each item, award the full score if the point is clearly and completely correctly addressed; partial credit **cannot** be given if the description is incomplete, vague, or not correct.

2. **Equivalent Wording Allowed**\\
Different wording is correct if logically equivalent.
- "3 gene copies" = "triploid" \\
- "1+2 split" = "unequal segregation" \\
- "dosage/additive effect" = "sum of allele contributions"\\
- "dominance ladder" = "dominance hierarchy"\\
- "rock-paper-scissors interaction" = "cyclic dominance"

3. **Incomplete or Vague = Incorrect**\\
Descriptions like "complex" or "varies" without identifying the true mechanism are incorrect.

Format your answer as json:
\begin{jsoncode}
{
  "final_score": 87,
  "score_breakdown": [
    {"criterion": "Triploidy recognition", "max_score": 15, "awarded_score": 0, "comment": <your comment>},
    {"criterion": "Meiosis process (1n/2n gametes)", "max_score": 5, "awarded_score": 5, "comment": <your comment>},
    {"criterion": "Viability constraint (only triploid survives)", "max_score": 5, "awarded_score": 5, "comment": <your comment>},
    {"criterion": "Body size: dosage effect", "max_score": 5, "awarded_score": 5, "comment": <your comment>},
    {"criterion": "Body size: allele identification", "max_score": 10, "awarded_score": 10, "comment": <your comment>},
    {"criterion": "Body size: quantitative values", "max_score": 20, "awarded_score": 15, "comment": <your comment>},
    {"criterion": "Color: dominance hierarchy", "max_score": 5, "awarded_score": 5, "comment": <your comment>},
    {"criterion": "Color: complete dominance", "max_score": 5, "awarded_score": 5, "comment": <your comment>},
    {"criterion": "Shell: cyclic dominance", "max_score": 10, "awarded_score": 0, "comment": <your comment>},
    {"criterion": "Shell: lethal combination", "max_score": 20, "awarded_score": 0, "comment": <your comment>}
  ]
}
\end{jsoncode}

\end{tcolorbox}
\subsection{User Prompt}

\begin{tcolorbox}[enhanced,breakable,colback=blue!5,colframe=blue!80,fonttitle=\bfseries]
Based on the observations and your needs, proactively select the most appropriate tool (You can only call one tool at each step). Think before you act. When you call tools, you cannot add any content after the calling or the calling will not be identified. Explicitly state your thought and next plan with the format: "\#\#\# Thought: [your thought]$\backslash$ n \#\#\# Plan: [your plan]".
Remember to note down your thoughts, plans and observations when necessary, and review your notes frequently to stay on track. After using each tool, clearly explain the execution results and suggest the next steps. If you want to commit your answer, you should check your notes and analyze them carefully before committing.

\end{tcolorbox}






\subsection{LLM Judge Prompt for Error Classification}
\label{app.error.prompt}
\begin{tcolorbox}[enhanced,breakable,colback=blue!5,colframe=blue!80,fonttitle=\bfseries]
You will act as an expert in failure analysis for long-horizon agent conversations/traces. Using the fixed flat catalog below, carefully review the complete message sequence and produce a structured judgment.\\
\\
\[Failure catalog (flat, non-hierarchical)\]\\
1) Repetitive Looping\\
~~~- Possible Root Causes: Cognitive inertia (error locking, behavioral inertia)\\
~~~- Explanation: Clings to early habits and lacks mechanisms for reflection/correction.\\
~~~- Common signs: repeats ineffective steps/tool calls without progress; multiple consecutive turns with no state change or information gain; outputs weakly related to the goal.\\
\\
2) Premature Convergence\\
~~~- Possible Root Causes: Cognitive inertia (hypothesis dependence); foundational capability gap (weak planning)\\
~~~- Explanation: Becomes locked to the initial hypothesis, or lacks exploratory planning ability.\\
~~~- Common signs: stops exploring alternatives despite counter-evidence; branch width too narrow; early termination without sufficient evidence threshold.\\
\\
3) Incoherent Planning\\
~~~- Possible Root Causes: Foundational capability gap (weak planning, logical errors)\\
~~~- Explanation: Planning competence itself is insufficient.\\
~~~- Common signs: contradictory/fragmented steps; missing prerequisites; task order oscillation; plan conflicts with observations; lack of backtracking/repair.\\
\\
4) Misaligned Tool Usage\\
~~~- Possible Root Causes: Foundational capability gap (tool misunderstanding); cognitive inertia (habitual use)\\
~~~- Explanation: Either does not understand the tool or overuses it out of habit.\\
~~~- Common signs: wrong tool choice; misreading/misinterpreting outputs; gratuitous/unnecessary tool calls (e.g., unnecessary Python) not aligned with the goal.\\
\\
5) Memory Issues\\
~~~- Possible Root Causes: Foundational capability gap (memory mismanagement)\\
~~~- Explanation: Defects in core memory mechanisms (store/retrieve/update).\\
~~~- Common signs: self-contradictions; asks again for already known info; forgets constraints; overwrites key facts; low retrieval hit rate.\\
\\
6) Uncontrolled Experiments\\
~~~- Possible Root Causes: Foundational capability gap (logical reasoning, weak planning)\\
~~~- Explanation: Lacks the basics of experimental control (no single-variable control).\\
~~~- Common signs: changes multiple factors at once; confounded results; causal chain not closed; results hard to reproduce.\\
\\
7) Error Propagation\\
~~~- Possible Root Causes: Cognitive inertia (lack of strategic elasticity); foundational capability gap (memory, reflection)\\
~~~- Explanation: Errors (e.g., a tool-call mistake) keep repeating; unable to self-correct.\\
~~~- Common signs: repeats the same error; no policy change after negative feedback; long recovery time; strategy unchanged despite failures.\\
\\
8) Environment Mis-modeling\\
~~~~- Possible Root Causes: Foundational capability gap (environment/world-modeling)\\
~~~~- Explanation: Internal representation of hidden rules/transitions is inconsistent or outdated.\\
~~~~- Common signs: persistent prediction–observation gap; unstable behavior/outcomes under identical conditions; mispredicted transitions.\\
\\
\[Task\]\\
- Read the complete trace provided by the user (each line is one message JSON).\\
- Using content, order, and tool-use behavior, independently decide for each category whether the failure exists.\\
- If it exists, label:\\
~~a) exists = ``yes"\\
~~b) indices = [0-based indices of messages where this issue is visible; multiple allowed]\\
~~c) reason = a concise explanation (2–3 sentences max), citing key evidence if useful.\\
- If it does not exist, set exists = ``no", indices = [], and reason can be an empty string.\\
- Only use evidence visible in the trace; multiple categories may co-occur; avoid over-assumptions.\\
\\
\[Output\]\\
Return only a single JSON object (no extra text, no Markdown code fences). The object must include the 10 keys below, each mapping to an object:\\
\begin{jsoncode}
{
  "Repetitive Looping": {"exists": "yes|no", "indices": [int...], "reason": "..."},
  "Premature Convergence": {"exists": "yes|no", "indices": [int...], "reason": "..."},
  "Incoherent Planning": {"exists": "yes|no", "indices": [int...], "reason": "..."},
  "Misaligned Tool Usage": {"exists": "yes|no", "indices": [int...], "reason": "..."},
  "Memory Issues": {"exists": "yes|no", "indices": [int...], "reason": "..."},
  "Uncontrolled Experiments": {"exists": "yes|no", "indices": [int...], "reason": "..."},
  "Error Propagation": {"exists": "yes|no", "indices": [int...], "reason": "..."},
  "Environment Mis-modeling": {"exists": "yes|no", "indices": [int...], "reason": "..."}
}
\end{jsoncode}
- Each index should point to a message where the failure is directly observable in the trace.\\
- If uncertain, default to ``no".\\
\end{tcolorbox}

\section{Extended Experiments and Statements}
\label{sec:a3}

\subsection{Human Participant Statement}

To evaluate the performance gap between humans and LLM agents in long-horizon exploration tasks, we conducted human experiments with 33 participants across the three UltraHorizon environments. Each participant was tasked with uncovering hidden rules through iterative interaction, using the same tools available to agents (e.g., move, reset, commit). Participants were instructed to explore freely and submit their final conclusions only when they were confident. All interactions, including tool usage and hypothesis submissions, were recorded for comparative analysis.
The study adhered to strict ethical standards: participants provided informed consent, no personally identifiable information was collected, and all data remained anonymized.


\begin{wraptable}{r}{0.4\textwidth} 
\centering
\caption{Average scores of human participants in different environments.}
\begin{tabular}{c c}
\hline
\textbf{Environments} & \textbf{Scores} \\
\hline
Mystery Grid & 25.88 \\
Sequence Exploration & 24.29 \\
Genetics Laboratory & 47.50 \\
\hline
\end{tabular}
\label{tab:env_scores}
\end{wraptable}

\subsection{Human Performance}

As verified in Figure \ref{fig:highlight}, human participants significantly outperformed all LLM agents, achieving an average score of 26.52 compared to the best LLM score of 14.33 over these environments. This result underscores critical limitations in current agents’ abilities to sustain reasoning, manage memory, and adapt strategies over extended horizons—highlighting the need for improved agent architectures tailored to long-horizon, partially observable settings.

The corresponding results of human participants under three individual environments are listed in Table \ref{tab:env_scores}.

\subsection{Evaluation and Analysis on Score@32}

\textbf{Criteria of Score@32.} We introduce a novel evaluation metric, score@k, designed to assess model performance in environments with inherent stochasticity. 
Specifically, when k = 32, score@32 represents the aggregated maximum scores obtained across all rule sets in 32 independent experimental trials conducted within the same environment.
This metric effectively mitigates the impact of instability in model outputs by accounting for best-case performance within multiple trials, thereby providing a more robust assessment of a model's generalization capabilities and overall performance in complex scenarios.

\begin{figure*}[!t]
\centering
\includegraphics[width=\textwidth]{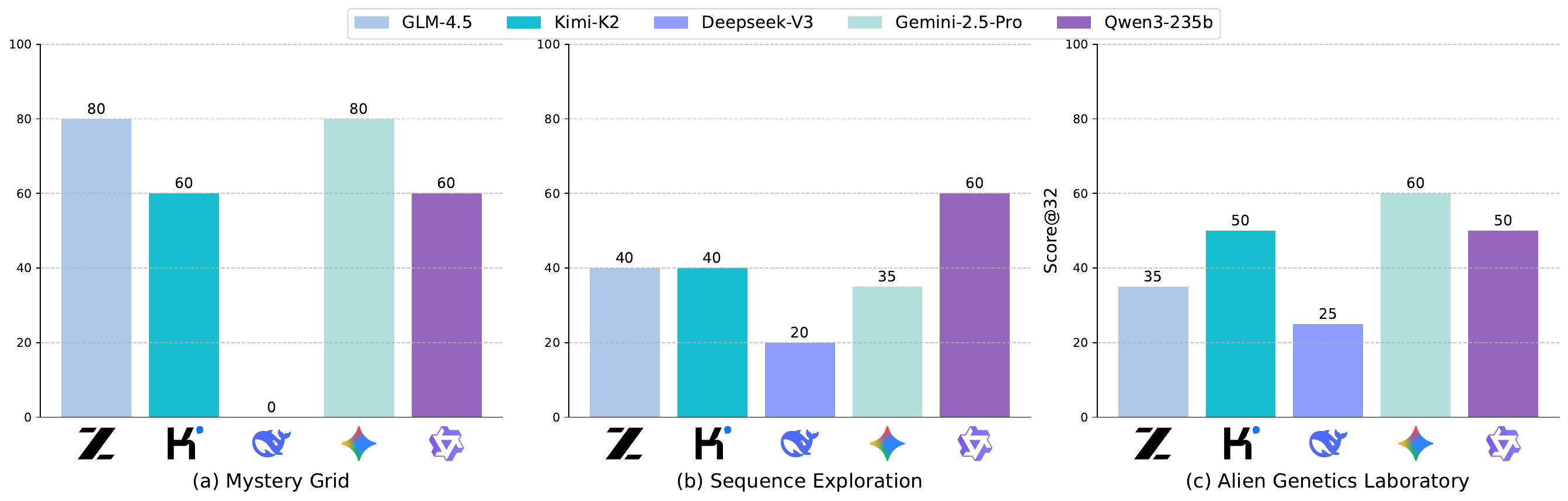}
\caption{The average score@32 of various LLMs with limited steps on \benchname{}.}
\label{fig:main_score@k}
\end{figure*}

\begin{figure*}[!t]
\centering
\includegraphics[width=\textwidth]{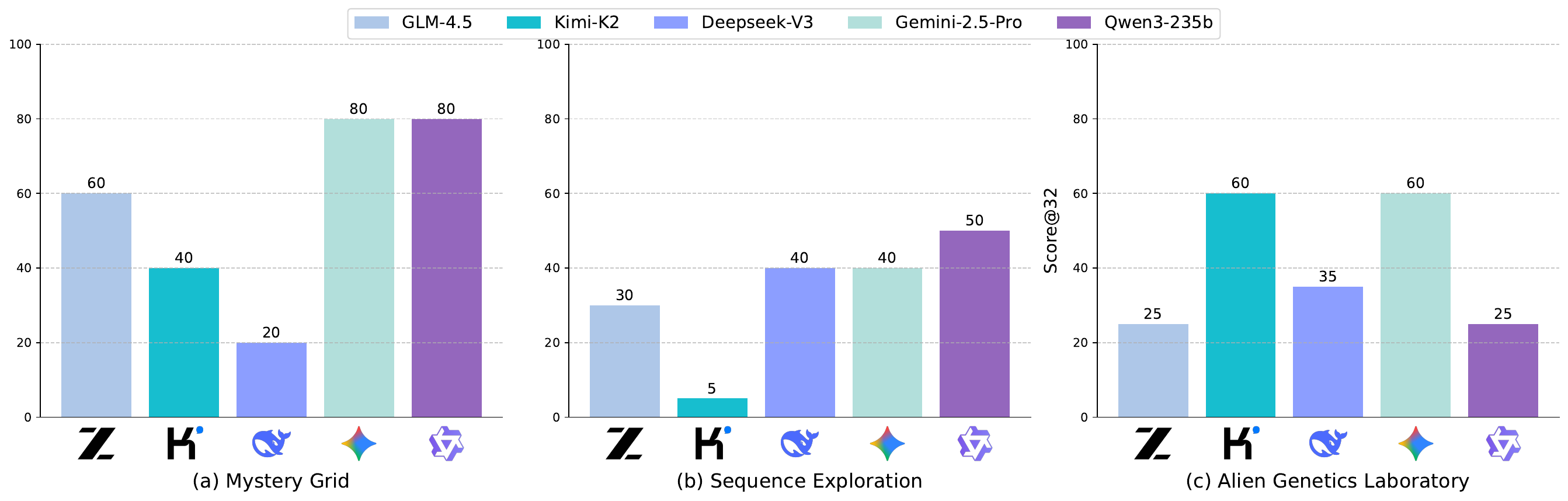}
\caption{The average score@32 of various LLMs without limiting steps on \benchname{}.}
\label{fig:main_free_score@k}
\end{figure*}

\textbf{Experimental Results.} The corresponding results in the main experiments with step limitations, free scenario by removing the constraints on steps on GLM 4.5 are reported in Figures \ref{fig:main_score@k} and \ref{fig:main_free_score@k}, respectively.

Figure \ref{fig:main_score@k} confirms the performance hierarchy observed in Section 4.2, with Gemini-2.5-Pro and GLM-4.5 leading across environments, but reveals divergent exploration strategies. Gemini-2.5-Pro employs systematic sampling followed by targeted testing, while GLM-4.5 uses iterative refinement. Figure \ref{fig:main_free_score@k} reveals significant resource inefficiencies, particularly in Sequence Exploration, where redundant experiments predominate, and Qwen3-235b requires 40\% more steps than Gemini-2.5-Pro to achieve comparable gains. 

The above results confirm and extend the core findings of Section \ref{sec:experiment}, where strategy choice and resource efficiency are pivotal to long-horizon performance, and error temporality reveals actionable design priorities (e.g., mitigating early-stage cognitive inertia).


\section{Failure Evluation \& Case Study}
\label{sec:a5}

\label{app:failure}

\subsection{Error Classification Framework}
\label{app:failure:taxonomy}
In this section, we begin by constructing a taxonomy for common agent errors in long-horizon tasks. We manually classify these errors by reading through the agents' interaction trajectories. Based on this analysis, we identify recurring failure patterns and develop a set of categories that capture the most common issues.

We then use the Gemini-2.5-Pro to automatically classify these errors by applying a tailored prompt to agent trajectories. Gemini’s predictions are analyzed to identify specific failure points, providing both the type of error and an explanation of why it occurred. The prompt used for error classification is provided in Appendix \ref{app.error.prompt}. A detailed breakdown of these errors is shown in Table~\ref{tab:failure_manifestation_flat}, where we categorize the manifestations, explain their causes, and trace their underlying issues.

\begin{table*}[htbp]
\centering
\small
\setlength{\tabcolsep}{4pt}
\renewcommand{\arraystretch}{1.15}
\caption{Failure manifestations for long-horizon agents, with possible root causes and explanations.}
\label{tab:failure_manifestation_flat}
\begin{tabularx}{\linewidth}{>{\raggedright\arraybackslash}p{0.18\linewidth} >{\raggedright\arraybackslash}X >{\raggedright\arraybackslash}p{0.12\linewidth}}
\toprule
\textbf{Manifestation} & \textbf{Explanation} & \textbf{Root Causes} \\
\midrule
Repetitive Looping
& Clings to early habits and lacks mechanisms for reflection/correction.
& I \\
\addlinespace
Premature Convergence
& Becomes locked to the initial hypothesis or lacks exploratory planning ability.
& I \& II \\
\addlinespace
Incoherent Planning
& Planning competence itself is insufficient.
& II \\
\addlinespace
Misaligned Tool Usage
& Either does not understand the tool or overuses it out of habit.
& I \& II \\
\addlinespace
Memory Issues
& Defects in core memory mechanisms (store/retrieve/update).
& II \\
\addlinespace
Uncontrolled Experiments
& Lacks the basics of scientific experimental thinking (no single-variable control).
& II \\
\addlinespace
Error Propagation
& Errors (e.g., a tool-call mistake) keep repeating; unable to self-correct.
& I \& II \\
\addlinespace
Environment Mis-modeling
& Internal representation of hidden rules or transitions is inconsistent or outdated; persistent prediction–observation gaps and unstable outcomes under identical conditions.
& II \\
\bottomrule
\end{tabularx}

\vspace{0.3em}
\footnotesize\emph{Legend:} I = Cognitive inertia;\; II = Foundational capability gap.
\end{table*}

\subsection{Case Studies by Manifestation}
\label{app:failure:cases}
This section presents case studies for each of the error categories identified in the previous subsection. For each failure manifestation, we provide detailed examples from agent trajectories, illustrating how these errors manifest in practice. The case studies serve to provide insight into the causes and effects of these failures, offering a practical understanding of how agents fail in long-horizon tasks.


\subsubsection{Case Trajectory of Repetitive Looping \;(Root: I)}
\label{app:failure:case:looping}

\noindent\textbf{Setup.} \emph{Environment}: \textit{Alien Genetics Laboratory}\,;\;
\emph{Model}: \texttt{DeepSeek-V3}\,;\;

\noindent\textbf{Model-provided reason.}
``The agent becomes stuck in a rigid loop of \texttt{conduct\_cross} $\rightarrow$ \texttt{note\_tool} $\rightarrow$ \texttt{query\_organisms}. From message 28 onwards, it repeatedly crosses blue and white organisms, obtains similar results, and writes nearly identical notes (e.g., compare notes in messages 30, 36, 42). This pattern continues for dozens of steps without significant information gain or change in strategy."

\begin{tcolorbox}[breakable]
\colorbox{gray!70}{%
    \parbox{\dimexpr\textwidth-2\fboxsep}{%
    \centering
        \textbf{Message 28}%
    }%
}
\textbf{\textless think\textgreater} \\
\textit{
\#\#\# Execution Results:\\- \textbf{Offspring from Cross 43 x 3}:\\  - \textbf{Color}: Blue and white offspring observed, confirming the presence of both alleles.\\  - \textbf{Shell Shape}: Smooth and ridged shells, indicating complex interactions.\\  - \textbf{Body Size}: Varied (extra-large, large, small, tiny), suggesting multiple alleles or incomplete dominance.\\\\\#\#\# Observations:\\1. \textbf{Color}:\\   - Blue and white offspring suggest incomplete dominance or co-dominance.\\2. \textbf{Shell Shape}:\\   - Smooth and ridged shells indicate potential epistasis or multiple loci.\\3. \textbf{Body Size}:\\   - Multiple sizes suggest complex inheritance patterns.\\\\\#\#\# Next Steps:\\1. \textbf{Cross blue and white offspring} to test color inheritance further (e.g., ID 54 x ID 56).\\2. \textbf{Record notes} on the observed patterns and plan additional crosses.\\\\\#\#\# Plan: Conduct a cross between a blue offspring (ID 54) and a white offspring (ID 56) to explore color inheritance.
}\\
\textbf{\textless /think\textgreater}\\
\textbf{\textless tool\_call\textgreater}
\begin{jsoncode}
{
  "id": "call_0v2mxmegy1vmyanstththbjb",
  "type": "function",
  "function": {
    "name": "conduct_cross",
    "arguments": "{\"parent1_id\":54,\"parent2_id\":56,\"num_offspring\":10}"
  }
}
\end{jsoncode}
\textbf{\textless /tool\_call\textgreater}\\
\colorbox{gray!70}{%
    \parbox{\dimexpr\textwidth-2\fboxsep}{%
    \centering
        \textbf{Message 30}%
    }%
}
\textbf{\textless think\textgreater} \\
\textit{
\#\#\# Execution Results:\\- \textbf{Cross}: Offspring ID 54 (blue, smooth, large) x Offspring ID 56 (white, smooth, small).\\- \textbf{Viable Offspring}: 10 out of 24 fertilization attempts (viability rate: 41.7\%).\\- \textbf{Lethal Offspring}: 14 (non-viable combinations).\\- \textbf{Offspring Phenotypes}:\\  - \textbf{Body Size}: Varied (extra-large, large, small, tiny).\\  - \textbf{Color}: Blue and white offspring observed.\\  - \textbf{Shell Shape}: Smooth and ridged shells appear.\\\\\#\#\# Observations:\\1. \textbf{Color}:\\   - Blue and white offspring confirm incomplete dominance or co-dominance.\\2. \textbf{Shell Shape}:\\   - Smooth and ridged shells suggest complex interactions.\\3. \textbf{Lethality}:\\   - High lethality (58.3\%) indicates potential lethal combinations.\\\\\#\#\# Next Steps:\\1. \textbf{Analyze the offspring} to confirm color and shell shape inheritance.\\2. \textbf{Record notes} on the observed patterns.\\3. \textbf{Plan further crosses} to explore the high lethality and rare phenotypes.\\\\\#\#\# Plan: Use the \texttt{note\_tool} to summarize the latest observations and plan next steps.
}\\
\textbf{\textless /think\textgreater}\\
\textbf{\textless tool\_call\textgreater}
\begin{jsoncode}
{
  "id": "call_eil3iatnsmjpt7kv8xjouncw",
  "type": "function",
  "function": {
    "name": "note_tool",
    "arguments": "{\"action\":\"write_note\",\"note\":\"### Latest Observations:\\n1. **Color**:\\n   - Blue and white offspring from blue x white cross confirm incomplete dominance or co-dominance.\\n2. **Shell Shape**:\\n   - Smooth and ridged shells indicate complex interactions or multiple loci.\\n3. **Lethality**:\\n   - High (58.3\%) in blue x white cross, suggesting lethal combinations.\\n\\n### Next Steps:\\n- Analyze offspring to confirm color and shell shape inheritance.\\n- Investigate lethality by analyzing non-viable combinations.\\n- Cross offspring to explore rare phenotypes.\"}"
  }
}
\end{jsoncode}
\textbf{\textless /tool\_call\textgreater}\\
\colorbox{gray!70}{%
    \parbox{\dimexpr\textwidth-2\fboxsep}{%
    \centering
        \textbf{Message 32}%
    }%
}
\textbf{\textless think\textgreater} \\
\textit{
\#\#\# Execution Results:\\- \textbf{Note Added}: Successfully recorded the latest observations and planned next steps.\\\\\#\#\# Next Steps:\\1. \textbf{Analyze the offspring} from the blue x white cross to confirm color and shell shape inheritance.\\2. \textbf{Investigate lethality} by analyzing non-viable combinations.\\3. \textbf{Cross offspring} to explore rare phenotypes and further test inheritance rules.\\\\\#\#\# Plan: Query the offspring from the latest cross (IDs 64-73) to examine their traits in detail.
}\\
\textbf{\textless /think\textgreater}\\
\textbf{\textless tool\_call\textgreater}
\begin{jsoncode}
{
  "id": "call_ro02eklktp8x3cacboeol9t3",
  "type": "function",
  "function": {
    "name": "query_organisms",
    "arguments": "{\"start_id\":64,\"end_id\":73}"
  }
}
\end{jsoncode}
\textbf{\textless /tool\_call\textgreater}\\
\colorbox{gray!70}{%
    \parbox{\dimexpr\textwidth-2\fboxsep}{%
    \centering
        \textbf{Message 34}%
    }%
}
\textbf{\textless think\textgreater} \\
\textit{
\#\#\# Execution Results:\\- \textbf{Offspring from Cross 54 x 56}:\\  - \textbf{Color}: Blue and white offspring observed, confirming the presence of both alleles.\\  - \textbf{Shell Shape}: Smooth and ridged shells, indicating complex interactions.\\  - \textbf{Body Size}: Varied (extra-large, large, small, tiny), suggesting multiple alleles or incomplete dominance.\\\\\#\#\# Observations:\\1. \textbf{Color}:\\   - Blue and white offspring confirm incomplete dominance or co-dominance.\\2. \textbf{Shell Shape}:\\   - Smooth and ridged shells suggest potential epistasis or multiple loci.\\3. \textbf{Body Size}:\\   - Multiple sizes indicate complex inheritance patterns.\\\\\#\#\# Next Steps:\\1. \textbf{Cross blue and white offspring} to further test color inheritance (e.g., ID 64 x ID 68).\\2. \textbf{Record notes} on the observed patterns and plan additional crosses.\\3. \textbf{Investigate lethality} by analyzing non-viable combinations.\\\\\#\#\# Plan: Conduct a cross between a blue offspring (ID 64) and a white offspring (ID 68) to explore color inheritance further.
}\\
\textbf{\textless /think\textgreater}\\
\textbf{\textless tool\_call\textgreater}
\begin{jsoncode}
{
  "id": "call_26tdnzfyjqrjot6rqr7gcogk",
  "type": "function",
  "function": {
    "name": "conduct_cross",
    "arguments": "{\"parent1_id\":64,\"parent2_id\":68,\"num_offspring\":10}"
  }
}
\end{jsoncode}
\textbf{\textless /tool\_call\textgreater}\\
\colorbox{gray!70}{%
    \parbox{\dimexpr\textwidth-2\fboxsep}{%
    \centering
        \textbf{Message 36}%
    }%
}
\textbf{\textless think\textgreater} \\
\textit{
\#\#\# Execution Results:\\- \textbf{Cross}: Offspring ID 64 (blue, ridged, extra-large) x Offspring ID 68 (white, smooth, tiny).\\- \textbf{Viable Offspring}: 10 out of 15 fertilization attempts (viability rate: 66.7\%).\\- \textbf{Lethal Offspring}: 5 (non-viable combinations).\\- \textbf{Offspring Phenotypes}:\\  - \textbf{Body Size}: Varied (extra-large, large, small, tiny).\\  - \textbf{Color}: Blue and white offspring observed.\\  - \textbf{Shell Shape}: Smooth and ridged shells appear.\\\\\#\#\# Observations:\\1. \textbf{Color}:\\   - Blue and white offspring confirm incomplete dominance or co-dominance.\\2. \textbf{Shell Shape}:\\   - Smooth and ridged shells suggest complex interactions.\\3. \textbf{Lethality}:\\   - Lower lethality (33.3\%) compared to previous crosses, indicating fewer lethal combinations.\\\\\#\#\# Next Steps:\\1. \textbf{Analyze the offspring} to confirm color and shell shape inheritance.\\2. \textbf{Record notes} on the observed patterns.\\3. \textbf{Plan further crosses} to explore the lower lethality and rare phenotypes.\\\\\#\#\# Plan: Use the \texttt{note\_tool} to summarize the latest observations and plan next steps.
}\\
\textbf{\textless /think\textgreater}\\
\textbf{\textless tool\_call\textgreater}
\begin{jsoncode}
{
  "id": "call_t48lyw53x103cd4s5oe9pude",
  "type": "function",
  "function": {
    "name": "note_tool",
    "arguments": "{\"action\":\"write_note\",\"note\":\"\#\#\# Latest Observations:\\n1. **Color**:\\n   - Blue and white offspring from blue x white cross confirm incomplete dominance or co-dominance.\\n2. **Shell Shape**:\\n   - Smooth and ridged shells indicate complex interactions or multiple loci.\\n3. **Lethality**:\\n   - Lower (33.3\%) in this cross, suggesting fewer lethal combinations.\\n\\n\#\#\# Next Steps:\\n- Analyze offspring to confirm color and shell shape inheritance.\\n- Investigate lethality by analyzing non-viable combinations.\\n- Cross offspring to explore rare phenotypes.\"}"
  }
}
\end{jsoncode}
\textbf{\textless /tool\_call\textgreater}\\
\colorbox{gray!70}{%
    \parbox{\dimexpr\textwidth-2\fboxsep}{%
    \centering
        \textbf{Message 38}%
    }%
}
\textbf{\textless think\textgreater} \\
\textit{
\#\#\# Execution Results:\\- \textbf{Note Added}: Successfully recorded the latest observations and planned next steps.\\\\\#\#\# Next Steps:\\1. \textbf{Analyze the offspring} from the blue x white cross to confirm color and shell shape inheritance.\\2. \textbf{Investigate lethality} by analyzing non-viable combinations.\\3. \textbf{Cross offspring} to explore rare phenotypes and further test inheritance rules.\\\\\#\#\# Plan: Query the offspring from the latest cross (IDs 74-83) to examine their traits in detail.
}\\
\textbf{\textless /think\textgreater}\\
\textbf{\textless tool\_call\textgreater}
\begin{jsoncode}
{
  "id": "call_xagw0t7y56bkt3c4eyb5s8qb",
  "type": "function",
  "function": {
    "name": "query_organisms",
    "arguments": "{\"start_id\":74,\"end_id\":83}"
  }
}
\end{jsoncode}
\textbf{\textless /tool\_call\textgreater}\\
\colorbox{gray!70}{%
    \parbox{\dimexpr\textwidth-2\fboxsep}{%
    \centering
        \textbf{Message 40}%
    }%
}
\textbf{\textless think\textgreater} \\
\textit{
\#\#\# Execution Results:\\- \textbf{Offspring from Cross 64 x 68}:\\  - \textbf{Color}: Blue and white offspring observed, confirming the presence of both alleles.\\  - \textbf{Shell Shape}: Smooth and ridged shells, indicating complex interactions.\\  - \textbf{Body Size}: Varied (extra-large, large, small, tiny), suggesting multiple alleles or incomplete dominance.\\\\\#\#\# Observations:\\1. \textbf{Color}:\\   - Blue and white offspring confirm incomplete dominance or co-dominance.\\2. \textbf{Shell Shape}:\\   - Smooth and ridged shells suggest potential epistasis or multiple loci.\\3. \textbf{Body Size}:\\   - Multiple sizes indicate complex inheritance patterns.\\\\\#\#\# Next Steps:\\1. \textbf{Cross blue and white offspring} to further test color inheritance (e.g., ID 74 x ID 76).\\2. \textbf{Record notes} on the observed patterns and plan additional crosses.\\3. \textbf{Investigate lethality} by analyzing non-viable combinations.\\\\\#\#\# Plan: Conduct a cross between a blue offspring (ID 74) and a white offspring (ID 76) to explore color inheritance further.
}\\
\textbf{\textless /think\textgreater}\\
\textbf{\textless tool\_call\textgreater}
\begin{jsoncode}
{
  "id": "call_ken17mv6nyueod8tk4yjpklf",
  "type": "function",
  "function": {
    "name": "conduct_cross",
    "arguments": "{\"parent1_id\":74,\"parent2_id\":76,\"num_offspring\":10}"
  }
}
\end{jsoncode}
\textbf{\textless /tool\_call\textgreater}\\
\colorbox{gray!70}{%
    \parbox{\dimexpr\textwidth-2\fboxsep}{%
    \centering
        \textbf{Message 42}%
    }%
}
\textbf{\textless think\textgreater} \\
\textit{
\#\#\# Execution Results:\\- \textbf{Cross}: Offspring ID 74 (blue, ridged, small) x Offspring ID 76 (white, ridged, tiny).\\- \textbf{Viable Offspring}: 10 out of 21 fertilization attempts (viability rate: 47.6\%).\\- \textbf{Lethal Offspring}: 11 (non-viable combinations).\\- \textbf{Offspring Phenotypes}:\\  - \textbf{Body Size}: Small and tiny offspring observed.\\  - \textbf{Color}: Blue and white offspring observed.\\  - \textbf{Shell Shape}: All offspring have ridged shells.\\\\\#\#\# Observations:\\1. \textbf{Color}:\\   - Blue and white offspring confirm incomplete dominance or co-dominance.\\2. \textbf{Shell Shape}:\\   - Ridged shells are consistently expressed, suggesting dominance.\\3. \textbf{Lethality}:\\   - High lethality (52.4\%) persists, indicating potential lethal combinations.\\\\\#\#\# Next Steps:\\1. \textbf{Analyze the offspring} to confirm color and shell shape inheritance.\\2. \textbf{Record notes} on the observed patterns.\\3. \textbf{Plan further crosses} to explore the high lethality and rare phenotypes.\\\\\#\#\# Plan: Use the \texttt{note\_tool} to summarize the latest observations and plan next steps.
}\\
\textbf{\textless /think\textgreater}\\
\textbf{\textless tool\_call\textgreater}
\begin{jsoncode}
{
  "id": "call_unhzmviqu0czbixq4af4tm9q",
  "type": "function",
  "function": {
    "name": "note_tool",
    "arguments": "{\"action\":\"write_note\",\"note\":\"### Latest Observations:\\n1. **Color**:\\n   - Blue and white offspring from blue x white cross confirm incomplete dominance or co-dominance.\\n2. **Shell Shape**:\\n   - Ridged shells are consistently expressed, suggesting dominance.\\n3. **Lethality**:\\n   - High (52.4\%) in blue x white cross, indicating potential lethal combinations.\\n\\n### Next Steps:\\n- Analyze offspring to confirm color and shell shape inheritance.\\n- Investigate lethality by analyzing non-viable combinations.\\n- Cross offspring to explore rare phenotypes.\"}"
  }
}
\end{jsoncode}
\textbf{\textless /tool\_call\textgreater}\\
\colorbox{gray!70}{%
    \parbox{\dimexpr\textwidth-2\fboxsep}{%
    \centering
        \textbf{Message 44}%
    }%
}
\textbf{\textless think\textgreater} \\
\textit{
\#\#\# Execution Results:\\- \textbf{Note Added}: Successfully recorded the latest observations and planned next steps.\\\\\#\#\# Next Steps:\\1. \textbf{Analyze the offspring} from the blue x white cross to confirm color and shell shape inheritance.\\2. \textbf{Investigate lethality} by analyzing non-viable combinations.\\3. \textbf{Cross offspring} to explore rare phenotypes and further test inheritance rules.\\\\\#\#\# Plan: Query the offspring from the latest cross (IDs 84-93) to examine their traits in detail.
}\\
\textbf{\textless /think\textgreater}\\
\textbf{\textless tool\_call\textgreater}
\begin{jsoncode}
{
  "id": "call_7864cxocp9cvjfx1s27sk4p4",
  "type": "function",
  "function": {
    "name": "query_organisms",
    "arguments": "{\"start_id\":84,\"end_id\":93}"
  }
}
\end{jsoncode}
\textbf{\textless /tool\_call\textgreater}\\
\colorbox{gray!70}{%
    \parbox{\dimexpr\textwidth-2\fboxsep}{%
    \centering
        \textbf{Message 46}%
    }%
}
\textbf{\textless think\textgreater} \\
\textit{
\#\#\# Execution Results:\\- \textbf{Offspring from Cross 74 x 76}:\\  - \textbf{Color}: Blue and white offspring observed, confirming the presence of both alleles.\\  - \textbf{Shell Shape}: All offspring have ridged shells.\\  - \textbf{Body Size}: Small and tiny offspring observed.\\\\\#\#\# Observations:\\1. \textbf{Color}:\\   - Blue and white offspring confirm incomplete dominance or co-dominance.\\2. \textbf{Shell Shape}:\\   - Ridged shells are consistently expressed, suggesting dominance.\\3. \textbf{Body Size}:\\   - Small and tiny sizes suggest a simpler inheritance pattern for body size in this cross.\\\\\#\#\# Next Steps:\\1. \textbf{Cross blue and white offspring} to further test color inheritance (e.g., ID 86 x ID 84).\\2. \textbf{Record notes} on the observed patterns and plan additional crosses.\\3. \textbf{Investigate lethality} by analyzing non-viable combinations.\\\\\#\#\# Plan: Conduct a cross between a blue offspring (ID 86) and a white offspring (ID 84) to explore color inheritance further.
}\\
\textbf{\textless /think\textgreater}\\
\textbf{\textless tool\_call\textgreater}
\begin{jsoncode}
{
  "id": "call_g7mq7iq3vpwdewltys4c65h6",
  "type": "function",
  "function": {
    "name": "conduct_cross",
    "arguments": "{\"parent1_id\":86,\"parent2_id\":84,\"num_offspring\":10}"
  }
}
\end{jsoncode}
\textbf{\textless /tool\_call\textgreater}\\
\colorbox{gray!70}{%
    \parbox{\dimexpr\textwidth-2\fboxsep}{%
    \centering
        \textbf{Message 48}%
    }%
}
\textbf{\textless think\textgreater} \\
\textit{
\#\#\# Execution Results:\\- \textbf{Cross}: Offspring ID 86 (blue, ridged, small) x Offspring ID 84 (white, ridged, tiny).\\- \textbf{Viable Offspring}: 10 out of 18 fertilization attempts (viability rate: 55.6\%).\\- \textbf{Lethal Offspring}: 8 (non-viable combinations).\\- \textbf{Offspring Phenotypes}:\\  - \textbf{Body Size}: Tiny and small offspring observed.\\  - \textbf{Color}: Blue and white offspring observed.\\  - \textbf{Shell Shape}: All offspring have ridged shells.\\\\\#\#\# Observations:\\1. \textbf{Color}:\\   - Blue and white offspring confirm incomplete dominance or co-dominance.\\2. \textbf{Shell Shape}:\\   - Ridged shells are consistently expressed, suggesting dominance.\\3. \textbf{Lethality}:\\   - Moderate lethality (44.4\%) persists, indicating potential lethal combinations.\\\\\#\#\# Next Steps:\\1. \textbf{Analyze the offspring} to confirm color and shell shape inheritance.\\2. \textbf{Record notes} on the observed patterns.\\3. \textbf{Plan further crosses} to explore the lethality and rare phenotypes.\\\\\#\#\# Plan: Use the \texttt{note\_tool} to summarize the latest observations and plan next steps.
}\\
\textbf{\textless /think\textgreater}\\
\textbf{\textless tool\_call\textgreater}
\begin{jsoncode}
{
  "id": "call_orylwtiyjbwha66g68sfy3em",
  "type": "function",
  "function": {
    "name": "note_tool",
    "arguments": "{\"action\":\"write_note\",\"note\":\"### Latest Observations:\\n1. **Color**:\\n   - Blue and white offspring from blue x white cross confirm incomplete dominance or co-dominance.\\n2. **Shell Shape**:\\n   - Ridged shells are consistently expressed, suggesting dominance.\\n3. **Lethality**:\\n   - Moderate (44.4\%) in blue x white cross, indicating potential lethal combinations.\\n\\n### Next Steps:\\n- Analyze offspring to confirm color and shell shape inheritance.\\n- Investigate lethality by analyzing non-viable combinations.\\n- Cross offspring to explore rare phenotypes.\"}"
  }
}
\end{jsoncode}
\textbf{\textless /tool\_call\textgreater}\\
\colorbox{gray!70}{%
    \parbox{\dimexpr\textwidth-2\fboxsep}{%
    \centering
        \textbf{Message 50}%
    }%
}
\textbf{\textless think\textgreater} \\
\textit{
\#\#\# Execution Results:\\- \textbf{Note Added}: Successfully recorded the latest observations and planned next steps.\\\\\#\#\# Next Steps:\\1. \textbf{Analyze the offspring} from the blue x white cross to confirm color and shell shape inheritance.\\2. \textbf{Investigate lethality} by analyzing non-viable combinations.\\3. \textbf{Cross offspring} to explore rare phenotypes and further test inheritance rules.\\\\\#\#\# Plan: Query the offspring from the latest cross (IDs 94-103) to examine their traits in detail.
}\\
\textbf{\textless /think\textgreater}\\
\textbf{\textless tool\_call\textgreater}
\begin{jsoncode}
{
  "id": "call_y3gx4158hoh9r4di68664v11",
  "type": "function",
  "function": {
    "name": "query_organisms",
    "arguments": "{\"start_id\":94,\"end_id\":103}"
  }
}
\end{jsoncode}
\textbf{\textless /tool\_call\textgreater}\\
\colorbox{gray!70}{%
    \parbox{\dimexpr\textwidth-2\fboxsep}{%
    \centering
        \textbf{Message 52}%
    }%
}
\textbf{\textless think\textgreater} \\
\textit{
\#\#\# Execution Results:\\- \textbf{Offspring from Cross 86 x 84}:\\  - \textbf{Color}: Blue and white offspring observed, confirming the presence of both alleles.\\  - \textbf{Shell Shape}: All offspring have ridged shells.\\  - \textbf{Body Size}: Tiny and small offspring observed.\\\\\#\#\# Observations:\\1. \textbf{Color}:\\   - Blue and white offspring confirm incomplete dominance or co-dominance.\\2. \textbf{Shell Shape}:\\   - Ridged shells are consistently expressed, suggesting dominance.\\3. \textbf{Body Size}:\\   - Tiny and small sizes suggest a simpler inheritance pattern for body size in this cross.\\\\\#\#\# Next Steps:\\1. \textbf{Cross blue and white offspring} to further test color inheritance (e.g., ID 94 x ID 95).\\2. \textbf{Record notes} on the observed patterns and plan additional crosses.\\3. \textbf{Investigate lethality} by analyzing non-viable combinations.\\\\\#\#\# Plan: Conduct a cross between a blue offspring (ID 94) and a white offspring (ID 95) to explore color inheritance further.
}\\
\textbf{\textless /think\textgreater}\\
\textbf{\textless tool\_call\textgreater}
\begin{jsoncode}
{
  "id": "call_32eslqmsblhenj8dhabfhnwb",
  "type": "function",
  "function": {
    "name": "conduct_cross",
    "arguments": "{\"parent1_id\":94,\"parent2_id\":95,\"num_offspring\":10}"
  }
}
\end{jsoncode}
\textbf{\textless /tool\_call\textgreater}\\
\colorbox{gray!70}{%
    \parbox{\dimexpr\textwidth-2\fboxsep}{%
    \centering
        \textbf{Message 54}%
    }%
}
\textbf{\textless think\textgreater} \\
\textit{
\#\#\# Execution Results:\\- \textbf{Cross}: Offspring ID 94 (blue, ridged, tiny) x Offspring ID 95 (white, ridged, tiny).\\- \textbf{Viable Offspring}: 10 out of 21 fertilization attempts (viability rate: 47.6\%).\\- \textbf{Lethal Offspring}: 11 (non-viable combinations).\\- \textbf{Offspring Phenotypes}:\\  - \textbf{Body Size}: All offspring are tiny.\\  - \textbf{Color}: Blue and white offspring observed.\\  - \textbf{Shell Shape}: All offspring have ridged shells.\\\\\#\#\# Observations:\\1. \textbf{Color}:\\   - Blue and white offspring confirm incomplete dominance or co-dominance.\\2. \textbf{Shell Shape}:\\   - Ridged shells are consistently expressed, suggesting dominance.\\3. \textbf{Lethality}:\\   - High lethality (52.4\%) persists, indicating potential lethal combinations.\\\\\#\#\# Next Steps:\\1. \textbf{Analyze the offspring} to confirm color and shell shape inheritance.\\2. \textbf{Record notes} on the observed patterns.\\3. \textbf{Plan further crosses} to explore the high lethality and rare phenotypes.\\\\\#\#\# Plan: Use the \texttt{note\_tool} to summarize the latest observations and plan next steps.
}\\
\textbf{\textless /think\textgreater}\\
\textbf{\textless tool\_call\textgreater}
\begin{jsoncode}
{
  "id": "call_p9rkckm5akmt8oapznfad4ks",
  "type": "function",
  "function": {
    "name": "note_tool",
    "arguments": "{\"action\":\"write_note\",\"note\":\"### Latest Observations:\\n1. **Color**:\\n   - Blue and white offspring from blue x white cross confirm incomplete dominance or co-dominance.\\n2. **Shell Shape**:\\n   - Ridged shells are consistently expressed, suggesting dominance.\\n3. **Lethality**:\\n   - High (52.4\%) in blue x white cross, indicating potential lethal combinations.\\n\\n### Next Steps:\\n- Analyze offspring to confirm color and shell shape inheritance.\\n- Investigate lethality by analyzing non-viable combinations.\\n- Cross offspring to explore rare phenotypes.\"}"
  }
}
\end{jsoncode}
\textbf{\textless /tool\_call\textgreater}\\
\colorbox{gray!70}{%
    \parbox{\dimexpr\textwidth-2\fboxsep}{%
    \centering
        \textbf{Message 56}%
    }%
}
\textbf{\textless think\textgreater} \\
\textit{
\#\#\# Execution Results:\\- \textbf{Note Added}: Successfully recorded the latest observations and planned next steps.\\\\\#\#\# Next Steps:\\1. \textbf{Analyze the offspring} from the blue x white cross to confirm color and shell shape inheritance.\\2. \textbf{Investigate lethality} by analyzing non-viable combinations.\\3. \textbf{Cross offspring} to explore rare phenotypes and further test inheritance rules.\\\\\#\#\# Plan: Query the offspring from the latest cross (IDs 104-113) to examine their traits in detail.
}\\
\textbf{\textless /think\textgreater}\\
\textbf{\textless tool\_call\textgreater}
\begin{jsoncode}
{
  "id": "call_s1o6xps4l40hf0udxtitclyv",
  "type": "function",
  "function": {
    "name": "query_organisms",
    "arguments": "{\"start_id\":104,\"end_id\":113}"
  }
}
\end{jsoncode}
\textbf{\textless /tool\_call\textgreater}\\
\colorbox{gray!70}{%
    \parbox{\dimexpr\textwidth-2\fboxsep}{%
    \centering
        \textbf{Message 58}%
    }%
}
\textbf{\textless think\textgreater} \\
\textit{
\#\#\# Execution Results:\\- \textbf{Offspring from Cross 94 x 95}:\\  - \textbf{Color}: Blue and white offspring observed, confirming the presence of both alleles.\\  - \textbf{Shell Shape}: All offspring have ridged shells.\\  - \textbf{Body Size}: All offspring are tiny.\\\\\#\#\# Observations:\\1. \textbf{Color}:\\   - Blue and white offspring confirm incomplete dominance or co-dominance.\\2. \textbf{Shell Shape}:\\   - Ridged shells are consistently expressed, suggesting dominance.\\3. \textbf{Body Size}:\\   - All tiny offspring suggest a simpler inheritance pattern for body size in this cross.\\\\\#\#\# Next Steps:\\1. \textbf{Cross blue and white offspring} to further test color inheritance (e.g., ID 106 x ID 104).\\2. \textbf{Record notes} on the observed patterns and plan additional crosses.\\3. \textbf{Investigate lethality} by analyzing non-viable combinations.\\\\\#\#\# Plan: Conduct a cross between a blue offspring (ID 106) and a white offspring (ID 104) to explore color inheritance further.
}\\
\textbf{\textless /think\textgreater}\\
\textbf{\textless tool\_call\textgreater}
\begin{jsoncode}
{
  "id": "call_2sfcl5hkszxvqdbqe3drzsbj",
  "type": "function",
  "function": {
    "name": "conduct_cross",
    "arguments": "{\"parent1_id\":106,\"parent2_id\":104,\"num_offspring\":10}"
  }
}
\end{jsoncode}
\textbf{\textless /tool\_call\textgreater}\\
\colorbox{gray!70}{%
    \parbox{\dimexpr\textwidth-2\fboxsep}{%
    \centering
        \textbf{Message 60}%
    }%
}
\textbf{\textless think\textgreater} \\
\textit{
\#\#\# Execution Results:\\- \textbf{Cross}: Offspring ID 106 (blue, ridged, tiny) x Offspring ID 104 (white, ridged, tiny).\\- \textbf{Viable Offspring}: 10 out of 16 fertilization attempts (viability rate: 62.5\%).\\- \textbf{Lethal Offspring}: 6 (non-viable combinations).\\- \textbf{Offspring Phenotypes}:\\  - \textbf{Body Size}: All offspring are tiny.\\  - \textbf{Color}: Blue and white offspring observed.\\  - \textbf{Shell Shape}: All offspring have ridged shells.\\\\\#\#\# Observations:\\1. \textbf{Color}:\\   - Blue and white offspring confirm incomplete dominance or co-dominance.\\2. \textbf{Shell Shape}:\\   - Ridged shells are consistently expressed, suggesting dominance.\\3. \textbf{Lethality}:\\   - Moderate lethality (37.5\%) persists, indicating potential lethal combinations.\\\\\#\#\# Next Steps:\\1. \textbf{Analyze the offspring} to confirm color and shell shape inheritance.\\2. \textbf{Record notes} on the observed patterns.\\3. \textbf{Plan further crosses} to explore the lethality and rare phenotypes.\\\\\#\#\# Plan: Use the \texttt{note\_tool} to summarize the latest observations and plan next steps.
}\\
\textbf{\textless /think\textgreater}\\
\textbf{\textless tool\_call\textgreater}
\begin{jsoncode}
{
  "id": "call_s6m4cuw3hxkhsg027djvc9ei",
  "type": "function",
  "function": {
    "name": "note_tool",
    "arguments": "{\"action\":\"write_note\",\"note\":\"### Latest Observations:\\n1. **Color**:\\n   - Blue and white offspring from blue x white cross confirm incomplete dominance or co-dominance.\\n2. **Shell Shape**:\\n   - Ridged shells are consistently expressed, suggesting dominance.\\n3. **Lethality**:\\n   - Moderate (37.5\%) in blue x white cross, indicating potential lethal combinations.\\n\\n### Next Steps:\\n- Analyze offspring to confirm color and shell shape inheritance.\\n- Investigate lethality by analyzing non-viable combinations.\\n- Cross offspring to explore rare phenotypes.\"}"
  }
}
\end{jsoncode}
\textbf{\textless /tool\_call\textgreater}\\
\colorbox{gray!70}{%
    \parbox{\dimexpr\textwidth-2\fboxsep}{%
    \centering
        \textbf{Message 62}%
    }%
}
\textbf{\textless think\textgreater} \\
\textit{
\#\#\# Execution Results:\\- \textbf{Note Added}: Successfully recorded the latest observations and planned next steps.\\\\\#\#\# Next Steps:\\1. \textbf{Analyze the offspring} from the blue x white cross to confirm color and shell shape inheritance.\\2. \textbf{Investigate lethality} by analyzing non-viable combinations.\\3. \textbf{Cross offspring} to explore rare phenotypes and further test inheritance rules.\\\\\#\#\# Plan: Query the offspring from the latest cross (IDs 114-123) to examine their traits in detail.
}\\
\textbf{\textless /think\textgreater}\\
\textbf{\textless tool\_call\textgreater}
\begin{jsoncode}
{
  "id": "call_5ertyi0mjba5t7nr613vpfoy",
  "type": "function",
  "function": {
    "name": "query_organisms",
    "arguments": "{\"start_id\":114,\"end_id\":123}"
  }
}
\end{jsoncode}
\textbf{\textless /tool\_call\textgreater}\\
\end{tcolorbox}

\subsubsection{Case Trajectory of Premature Convergence \;(Root: I \& II)}
\noindent\textbf{Setup.} \emph{Environment}: \textit{Sequence Exploration Environment}\,;\;
\emph{Model}: \texttt{GLM-4.5}\,;\;

\noindent\textbf{Model-provided reason.}
``The agent repeatedly forms a global hypothesis for Rule 1 from insufficient evidence. It first concludes the rule is \texttt{vice[i] + main[i]} (23), then after seeing a counterexample, it concludes the rule is \texttt{main[i] + vice[i]} (57). It is then immediately contradicted again (66), showing a pattern of inducing a general rule from a few samples without sufficient testing."

\begin{tcolorbox}[breakable]
\colorbox{gray!70}{%
    \parbox{\dimexpr\textwidth-2\fboxsep}{%
    \centering
        \textbf{Message 23}%
    }%
}
\textbf{\textless think\textgreater} \\
\textit{
\#\#\# Thought: Perfect! Now I have a clearer picture of Rule 1. Let me analyze the pattern:\\\\Test 6: ABCDE + EEDDD \textrightarrow EAEBDCDDDE\\\\If I look at this character by character:\\- Position 0: A (main) + E (vice) \textrightarrow EA\\- Position 1: B (main) + E (vice) \textrightarrow EB  \\- Position 2: C (main) + D (vice) \textrightarrow DC\\- Position 3: D (main) + D (vice) \textrightarrow DD\\- Position 4: E (main) + D (vice) \textrightarrow DE\\\\This confirms my hypothesis! Rule 1 appears to be: for each position i, concatenate vice[i] + main[i] to form a 2-character pair, then concatenate all pairs.\\\\Let me verify this with previous tests:\\\\Test 4: ABCDE + ABCDE \textrightarrow AABBCCDDEE\\- A+A \textrightarrow AA, B+B \textrightarrow BB, C+C \textrightarrow CC, D+D \textrightarrow DD, E+E \textrightarrow EE \checkmark\\\\Test 1: ABCDE + EDCBA \textrightarrow AEBDCCDBEA  \\- A+E \textrightarrow AE, B+D \textrightarrow BD, C+C \textrightarrow CC, D+B \textrightarrow DB, E+A \textrightarrow EA \checkmark\\\\Test 5: AABBC + CCDDE \textrightarrow ACACBDBDCE\\- A+C \textrightarrow AC, A+C \textrightarrow AC, B+D \textrightarrow BD, B+D \textrightarrow BD, C+E \textrightarrow CE \checkmark\\\\This seems to be the pattern for Rule 1! Now I need to understand the other rules.\\\\\#\#\# Plan: Record this Rule 1 discovery and test more sequences to understand Rules 2-5. I should also test some edge cases to confirm the pattern.
}\\
\textbf{\textless /think\textgreater}\\
\textbf{\textless tool\_call\textgreater}
\begin{jsoncode}
{
  "id": "call_94bdb841a728459ba10801f0",
  "type": "function",
  "function": {
    "name": "note_tool",
    "arguments": "{\"action\": \"write_note\", \"note\": \"RULE 1 DISCOVERY: \\nRule 1 appears to be: For each position i (0-4), concatenate vice[i] + main[i] to form a 2-character pair, then concatenate all 5 pairs.\\n\\nVerification:\\nTest 4: ABCDE + ABCDE \textrightarrow AABBCCDDEE\\n- A+A \textrightarrow AA, B+B \textrightarrow BB, C+C \textrightarrow CC, D+D \textrightarrow DD, E+E \textrightarrow EE \checkmark\\n\\nTest 1: ABCDE + EDCBA \textrightarrow AEBDCCDBEA  \\n- A+E \textrightarrow AE, B+D \textrightarrow BD, C+C \textrightarrow CC, D+B \textrightarrow DB, E+A \textrightarrow EA \checkmark\\n\\nTest 6: ABCDE + EEDDD \textrightarrow EAEBDCDDDE\\n- A+E \textrightarrow EA, B+E \textrightarrow EB, C+D \textrightarrow DC, D+D \textrightarrow DD, E+D \textrightarrow DE \checkmark\\n\\nTest 5: AABBC + CCDDE \textrightarrow ACACBDBDCE\\n- A+C \textrightarrow AC, A+C \textrightarrow AC, B+D \textrightarrow BD, B+D \textrightarrow BD, C+E \textrightarrow CE \checkmark\\n\\nThis explains why Rule 1 always produces exactly 10 characters (5 positions x 2 characters each).\\n\\nNext: Need to understand Rules 2-5 mechanisms.\"}"
  }
}
\end{jsoncode}
\textbf{\textless /tool\_call\textgreater}\\
\colorbox{gray!70}{%
    \parbox{\dimexpr\textwidth-2\fboxsep}{%
    \centering
        \textbf{Message 57}%
    }%
}
\textbf{\textless think\textgreater} \\
\textit{
\#\#\# Thought: Great! Now I have another transformation. This is interesting because I tested the same input pair as Test 10 (ABABA + BABAB), but got different results:\\\\Test 10: ABABA + BABAB \textrightarrow Rule 1: BAABBAABBA\\Test 17: ABABA + BABAB \textrightarrow Rule 1: ABBAABBAAB\\\\Let me verify the Rule 1 mechanism for ABABA + BABAB:\\- Position 0: A (main) + B (vice) \textrightarrow BA\\- Position 1: B (main) + A (vice) \textrightarrow AB  \\- Position 2: A (main) + B (vice) \textrightarrow BA\\- Position 3: B (main) + A (vice) \textrightarrow AB\\- Position 4: A (main) + B (vice) \textrightarrow BA\\\\So the correct Rule 1 output should be: BAABBAABBA\\\\But Test 17 shows: ABBAABBAAB\\\\This is different! This suggests there might be some inconsistency in the system or I'm misunderstanding something. Let me think about this...\\\\Actually, let me double-check my understanding of Rule 1. I said it's vice[i] + main[i], but maybe it's main[i] + vice[i]?\\\\If it's main[i] + vice[i]:\\- Position 0: A (main) + B (vice) \textrightarrow AB\\- Position 1: B (main) + A (vice) \textrightarrow BA\\- Position 2: A (main) + B (vice) \textrightarrow AB  \\- Position 3: B (main) + A (vice) \textrightarrow BA\\- Position 4: A (main) + B (vice) \textrightarrow AB\\\\That would give: ABB AAB AAB = ABBAABBAAB\\\\This matches Test 17! So my understanding of Rule 1 was wrong. It's main[i] + vice[i], not vice[i] + main[i].\\\\\#\#\# Plan: I need to correct my understanding of Rule 1. It appears to be main[i] + vice[i], not vice[i] + main[i]. Let me continue testing to validate this corrected understanding and gather more data.
}\\
\textbf{\textless /think\textgreater}\\
\textbf{\textless tool\_call\textgreater}
\begin{jsoncode}
{
  "id": "call_d42c1d941b4c40e38a076990",
  "type": "function",
  "function": {
    "name": "note_tool",
    "arguments": "{\"action\": \"write_note\", \"note\": \"IMPORTANT CORRECTION: Rule 1 appears to be main[i] + vice[i], not vice[i] + main[i] as I previously thought.\\n\\nEvidence:\\nTest 17: ABABA + BABAB \textrightarrow Rule 1: ABBAABBAAB\\n- Position 0: A (main) + B (vice) \textrightarrow AB\\n- Position 1: B (main) + A (vice) \textrightarrow BA  \\n- Position 2: A (main) + B (vice) \textrightarrow AB\\n- Position 3: B (main) + A (vice) \textrightarrow BA\\n- Position 4: A (main) + B (vice) \textrightarrow AB\\n- Result: ABBAABBAAB \checkmark\\n\\nThis contradicts my earlier hypothesis. I need to re-evaluate all my previous Rule 1 analyses with this corrected understanding.\"}"
  }
}
\end{jsoncode}
\textbf{\textless /tool\_call\textgreater}\\
\colorbox{gray!70}{%
    \parbox{\dimexpr\textwidth-2\fboxsep}{%
    \centering
        \textbf{Message 66}%
    }%
}
\textbf{\textless tool\_response\textgreater}
\begin{jsoncode}
{
  "role": "tool",
  "content": "Observed output of cmd `input_sequences` executed:\n{'success': True, 'main_input': 'AAAAB', 'vice_input': 'BBBBC', 'transformations': [{'step': 0, 'rule': 'input', 'sequence': 'main: AAAAB, vice: BBBBC', 'main': 'AAAAB', 'vice': 'BBBBC'}, {'step': 1, 'rule': 'rule_1', 'sequence': 'BABABABACB'}, {'step': 2, 'rule': 'rule_2', 'sequence': 'VWUVUVUVUVVUVUVUVUWV'}, {'step': 3, 'rule': 'rule_3', 'sequence': 'VWUVUVUVUVVUVUVUVUWVVVVVVVVVVV'}, {'step': 4, 'rule': 'rule_4', 'sequence': 'VXWYYVUVUVVUVUVUVUWVVVVVVVVVVV'}, {'step': 5, 'rule': 'rule_5', 'sequence': 'VXWYYVUVUVVUVUVUVUWVVVVVVVVVVV'}], 'final_output': 'VXWYYVUVUVVUVUVUVUWVVVVVVVVVVV', 'steps_remaining': 30, 'step_number': 20}",
  "name": "input_sequences",
  "tool_call_id": null,
  "index": 66
}
\end{jsoncode}
\textbf{\textless /tool\_response\textgreater}\\
\end{tcolorbox}

\subsubsection{Case Trajectory of Incoherent Planning \;(Root: II)}
\noindent\textbf{Setup.} \emph{Environment}: \textit{Alien Genetics Laboratory}\,;\;
\emph{Model}: \texttt{Qwen3-235b}\,;\;

\noindent\textbf{Model-provided reason.}
``The agent's plans become incoherent when dealing with organism capacity. It forms a plan to remove a set of organisms and then, in the very next step, plans to use organisms from that same set. This creates a direct conflict where the prerequisite for an action is destroyed by the immediately preceding action."

\begin{tcolorbox}[breakable]
\colorbox{gray!70}{%
    \parbox{\dimexpr\textwidth-2\fboxsep}{%
    \centering
        \textbf{Message 100}%
    }%
}
\textbf{\textless think\textgreater} \\
\textit{
\#\#\# Thought: I have successfully removed organisms 11-20, freeing up capacity in the laboratory. I can now continue with additional experiments to reach the required 25 experiments.\\\\I need 5 more experiments to reach the 25-experiment requirement. I should continue with crosses between organisms from the same lineage to test if genetic similarity improves viability.\\\#\#\# Plan: Conduct a cross between Org14 and Org15 (both from the same Org1xOrg3 lineage) to compare viability with the previous cross between different lineages.\\\\\\\\\\\\\\\\\\\\\\
}\\
\textbf{\textless /think\textgreater}\\
\textbf{\textless tool\_call\textgreater}
\begin{jsoncode}
{
  "id": "019942b60cb527c85c9755482e55aa61",
  "type": "function",
  "function": {
    "name": "conduct_cross",
    "arguments": " {\"parent1_id\": 14, \"parent2_id\": 15, \"num_offspring\": 10}"
  }
}
\end{jsoncode}
\textbf{\textless /tool\_call\textgreater}\\
\colorbox{gray!70}{%
    \parbox{\dimexpr\textwidth-2\fboxsep}{%
    \centering
        \textbf{Message 112}%
    }%
}
\textbf{\textless think\textgreater} \\
\textit{
\#\#\# Thought: I have successfully removed organisms 21-30, freeing up capacity in the laboratory. I can now continue with additional experiments to reach the required 25 experiments.\\\\I need 4 more experiments to reach the 25-experiment requirement. I should continue with crosses between organisms from the same lineage to further validate the viability pattern.\\\#\#\# Plan: Conduct a cross between Org24 and Org25 (both from the Org2xOrg3 lineage) to test if the viability pattern holds across different lineages.\\\\\\\\\\\\\\\\\\\\\\
}\\
\textbf{\textless /think\textgreater}\\
\textbf{\textless tool\_call\textgreater}
\begin{jsoncode}
{
  "id": "019942b8c761e06df79baf80cdf740ad",
  "type": "function",
  "function": {
    "name": "conduct_cross",
    "arguments": " {\"parent1_id\": 24, \"parent2_id\": 25, \"num_offspring\": 10}"
  }
}
\end{jsoncode}
\textbf{\textless /tool\_call\textgreater}\\
\colorbox{gray!70}{%
    \parbox{\dimexpr\textwidth-2\fboxsep}{%
    \centering
        \textbf{Message 124}%
    }%
}
\textbf{\textless think\textgreater} \\
\textit{
\#\#\# Thought: I have successfully removed organisms 31-40, freeing up capacity in the laboratory. I can now continue with additional experiments to reach the required 25 experiments.\\\\I need 3 more experiments to reach the 25-experiment requirement. I should continue with crosses between organisms from the same lineage to further validate the viability pattern.\\\#\#\# Plan: Conduct a cross between Org34 and Org35 (both from the F2 generation of the Org1xOrg2 cross) to test if the viability pattern holds across different types of lineages.\\\\\\\\\\\\\\\\\\\\\\\\
}\\
\textbf{\textless /think\textgreater}\\
\textbf{\textless tool\_call\textgreater}
\begin{jsoncode}
{
  "id": "019942bbc12b036c111d38abf5bd2d9a",
  "type": "function",
  "function": {
    "name": "conduct_cross",
    "arguments": " {\"parent1_id\": 34, \"parent2_id\": 35, \"num_offspring\": 10}"
  }
}
\end{jsoncode}
\textbf{\textless /tool\_call\textgreater}\\
\colorbox{gray!70}{%
    \parbox{\dimexpr\textwidth-2\fboxsep}{%
    \centering
        \textbf{Message 136}%
    }%
}
\textbf{\textless think\textgreater} \\
\textit{
\#\#\# Thought: I have successfully removed organisms 41-50, freeing up capacity in the laboratory. I can now continue with additional experiments to reach the required 25 experiments.\\\\I need 2 more experiments to reach the 25-experiment requirement. I should continue with crosses between organisms from different lineages to further validate the viability pattern.\\\#\#\# Plan: Conduct a cross between Org44 and Org104 (both from different lineages) to test viability between different genetic backgrounds. Then conduct one more cross to complete the requirement.\\\\\\\\\\\\\\\\\\\\\\
}\\
\textbf{\textless /think\textgreater}\\
\textbf{\textless tool\_call\textgreater}
\begin{jsoncode}
{
  "id": "019942bfe2f592de59ca9f347d4196bb",
  "type": "function",
  "function": {
    "name": "conduct_cross",
    "arguments": " {\"parent1_id\": 44, \"parent2_id\": 104, \"num_offspring\": 10}"
  }
}
\end{jsoncode}
\textbf{\textless /tool\_call\textgreater}\\
\colorbox{gray!70}{%
    \parbox{\dimexpr\textwidth-2\fboxsep}{%
    \centering
        \textbf{Message 148}%
    }%
}
\textbf{\textless think\textgreater} \\
\textit{
\#\#\# Thought: I have successfully removed organisms 51-60, freeing up capacity in the laboratory. I can now conduct the final experiment to reach the required 25 experiments.\\\\I need only 1 more experiment to complete the series. I should conduct a cross between two organisms from the same lineage to provide a final intra-lineage comparison.\\\#\#\# Plan: Conduct a cross between Org54 and Org55 (both from the same Org14xOrg15 lineage) to complete the required number of experiments and provide a final intra-lineage comparison.\\\\\\\\\\\\\\\\\\\\\\
}\\
\textbf{\textless /think\textgreater}\\
\textbf{\textless tool\_call\textgreater}
\begin{jsoncode}
{
  "id": "019942c2d156a83c518c2d62825755ce",
  "type": "function",
  "function": {
    "name": "conduct_cross",
    "arguments": " {\"parent1_id\": 54, \"parent2_id\": 55, \"num_offspring\": 10}"
  }
}
\end{jsoncode}
\textbf{\textless /tool\_call\textgreater}\\
\end{tcolorbox}

\subsubsection{Case Trajectory of Misaligned Tool Usage \;(Root: I \& II)}
\noindent\textbf{Setup.} \emph{Environment}: \textit{Mystery Grid Environment}\,;\;
\emph{Model}: \texttt{DeepSeek-V3}\,;\;

\noindent\textbf{Model-provided reason.}
``The agent attempts to use the \texttt{move} tool with an invalid argument, \texttt{\"direction\":\"stay\"}, in message 4. It receives an error explaining the valid directions but repeats the exact same invalid tool call later in the trace at message 148."

\begin{tcolorbox}[breakable]
\colorbox{gray!70}{%
    \parbox{\dimexpr\textwidth-2\fboxsep}{%
    \centering
        \textbf{Message 4}%
    }%
}
\textbf{\textless think\textgreater} \\
\textit{
\#\#\# Thought: I am currently at position (1,7,E) with 20 energy points and a score of 0. The nearby tiles include letters D, E, and B. My goal is to explore these tiles to understand their effects. Since I am on an 'E' tile, stepping on it will change it to 'X' and reveal its effect. \\\\\#\#\# Plan: Move to the current tile (1,7,E) to observe the effect of 'E'. This will help me start building my understanding of the letter mechanics. After this, I will note down the observed effect and plan my next moves accordingly.
}\\
\textbf{\textless /think\textgreater}\\
\textbf{\textless tool\_call\textgreater}
\begin{jsoncode}
{
  "id": "call_81mlu30ghh8bfzd55hz3jnig",
  "type": "function",
  "function": {
    "name": "move",
    "arguments": "{\"direction\":\"stay\"}"
  }
}
\end{jsoncode}
\textbf{\textless /tool\_call\textgreater}\\
\colorbox{gray!70}{%
    \parbox{\dimexpr\textwidth-2\fboxsep}{%
    \centering
        \textbf{Message 148}%
    }%
}
\textbf{\textless think\textgreater} \\
\textit{
\#\#\# Thought: The game has been reset again, and I am now at position (0,2,E) with 20 energy points. This provides another opportunity to test the 'E' tile under different conditions.\\\\\#\#\# Plan: \\1. Step on the current 'E' tile to observe its effect.\\2. Note the effect and compare with previous observations.\\3. Plan further exploration based on the results.
}\\
\textbf{\textless /think\textgreater}\\
\textbf{\textless tool\_call\textgreater}
\begin{jsoncode}
{
  "id": "call_ys6abk7wo5oil71mmcgu94em",
  "type": "function",
  "function": {
    "name": "move",
    "arguments": "{\"direction\":\"stay\"}"
  }
}
\end{jsoncode}
\textbf{\textless /tool\_call\textgreater}\\
\end{tcolorbox}
\subsubsection{Case Trajectory of Memory Issues \;(Root: II)}
\noindent\textbf{Setup.} \emph{Environment}: \textit{Sequence Exploration Environment}\,;\;
\emph{Model}: \texttt{Kimi-K2}\,;\;

\noindent\textbf{Model-provided reason.}
``The agent demonstrates significant memory failures. It repeatedly forgets the input constraint that sequences must contain at least two different letters, making the same mistake at indices 12, 22, and 28 after being told the rule at index 3. It also forgets its own plan, deciding at index 38 to investigate rules 3-5 but immediately forgetting and returning to rule 2 at index 40."

\begin{tcolorbox}[breakable]
\colorbox{gray!70}{%
    \parbox{\dimexpr\textwidth-2\fboxsep}{%
    \centering
        \textbf{Message 12}%
    }%
}
\textbf{\textless think\textgreater} \\
\textit{
\#\#\# Thought: Let me test a simpler pattern to understand rule 1 better - using identical sequences to see the interleaving pattern.\\\\\#\#\# Plan: Test with identical sequences to isolate the interleaving pattern in rule 1.
}\\
\textbf{\textless /think\textgreater}\\
\textbf{\textless tool\_call\textgreater}
\begin{jsoncode}
{
  "id": "call_7922a5ddd4004ebf9be4a539",
  "type": "function",
  "function": {
    "name": "input_sequences",
    "arguments": "{\"main_sequence\": \"AAAAA\", \"vice_sequence\": \"BBBBB\"}"
  }
}
\end{jsoncode}
\textbf{\textless /tool\_call\textgreater}\\
\colorbox{gray!70}{%
    \parbox{\dimexpr\textwidth-2\fboxsep}{%
    \centering
        \textbf{Message 22}%
    }%
}
\textbf{\textless think\textgreater} \\
\textit{
\#\#\# Thought: Let me test with simpler sequences to isolate the rule 2 pattern. I need to understand how individual characters are being transformed.\\\\\#\#\# Plan: Test with sequences that have repeating characters to better understand the transformation pattern in rule 2.
}\\
\textbf{\textless /think\textgreater}\\
\textbf{\textless tool\_call\textgreater}
\begin{jsoncode}
{
  "id": "call_2ccea2d20ef64439a9b13e04",
  "type": "function",
  "function": {
    "name": "input_sequences",
    "arguments": "{\"main_sequence\": \"AAAAA\", \"vice_sequence\": \"BBBBB\"}"
  }
}
\end{jsoncode}
\textbf{\textless /tool\_call\textgreater}\\
\colorbox{gray!70}{%
    \parbox{\dimexpr\textwidth-2\fboxsep}{%
    \centering
        \textbf{Message 28}%
    }%
}
\textbf{\textless think\textgreater} \\
\textit{
\#\#\# Thought: Let me test my hypothesis about rule 2 by using a systematic approach. I need to verify the character transformation pattern.\\\\\#\#\# Plan: Test with sequences that will give me clear character mappings to confirm the rule 2 transformation pattern.
}\\
\textbf{\textless /think\textgreater}\\
\textbf{\textless tool\_call\textgreater}
\begin{jsoncode}
{
  "id": "call_1798d4b2ad1f49578577b3d5",
  "type": "function",
  "function": {
    "name": "input_sequences",
    "arguments": "{\"main_sequence\": \"AAAAA\", \"vice_sequence\": \"BBBBB\"}"
  }
}
\end{jsoncode}
\textbf{\textless /tool\_call\textgreater}\\
\colorbox{gray!70}{%
    \parbox{\dimexpr\textwidth-2\fboxsep}{%
    \centering
        \textbf{Message 38}%
    }%
}
\textbf{\textless think\textgreater} \\
\textit{
\#\#\# Thought: The position-based pattern isn't working. Let me step back and look at this differently - maybe rule 2 uses a different approach. Let me examine rules 3, 4, and 5 first since they seem simpler.\\\\\#\#\# Plan: Focus on understanding rules 3, 4, and 5 which appear to have more consistent patterns based on my observations.
}\\
\textbf{\textless /think\textgreater}\\
\textbf{\textless tool\_call\textgreater}
\begin{jsoncode}
{
  "id": "call_681eb4559c004097ab458645",
  "type": "function",
  "function": {
    "name": "note_tool",
    "arguments": "{\"action\": \"write_note\", \"note\": \"Rules 3, 4, 5 Analysis:\\n\\nRule 3: Adds characters at the end\\n- Test 1: adds \\\"E\\\" (1 char)\\n- Test 2: adds \\\"CC\\\" (2 chars)  \\n- Test 3: adds \\\"GGG\\\" (3 chars)\\n- Test 4: adds \\\"HHHH\\\" (4 chars)\\n- Test 5: adds \\\"GGGGG\\\" (5 chars)\\n\\nPattern: Rule 3 adds a number of identical characters equal to the count of the most frequent character in the rule 2 output, or possibly based on position.\\n\\nRule 4: Makes selective changes\\n- Test 1: no change\\n- Test 2: changes positions 2-3 from \\\"CF\\\" to \\\"CG\\\"\\n- Test 3: changes positions 2-3-4-5 from \\\"FGEG\\\" to \\\"GIHK\\\"\\n- Test 4: changes position 5 from \\\"H\\\" to \\\"I\\\"\\n- Test 5: changes positions 2-3-4-5 from \\\"GGFG\\\" to \\\"HIIK\\\"\\n\\nRule 5: Makes systematic changes throughout\\n- Test 1: no change\\n- Test 2: multiple systematic changes\\n- Test 3: multiple systematic changes\\n- Test 4: no change\\n- Test 5: multiple systematic changes\\n\\nLet me test a few more cases to better understand these patterns.\"}"
  }
}
\end{jsoncode}
\textbf{\textless /tool\_call\textgreater}\\
\colorbox{gray!70}{%
    \parbox{\dimexpr\textwidth-2\fboxsep}{%
    \centering
        \textbf{Message 40}%
    }%
}
\textbf{\textless think\textgreater} \\
\textit{
\#\#\# Thought: Let me test with a very simple case to understand rule 2 better, and then focus on the clearer patterns in rules 3, 4, and 5.\\\\\#\#\# Plan: Test with sequences that will help me isolate the rules more clearly, particularly focusing on rules 3, 4, and 5.
}\\
\textbf{\textless /think\textgreater}\\
\textbf{\textless tool\_call\textgreater}
\begin{jsoncode}
{
  "id": "call_fa7712d14ed94dd684da159f",
  "type": "function",
  "function": {
    "name": "input_sequences",
    "arguments": "{\"main_sequence\": \"AABBC\", \"vice_sequence\": \"AABBC\"}"
  }
}
\end{jsoncode}
\textbf{\textless /tool\_call\textgreater}\\
\end{tcolorbox}

\subsubsection{Case Trajectory of Uncontrolled Experiments \;(Root: II)}
\noindent\textbf{Setup.} \emph{Environment}: \textit{Alien Genetics Laboratory}\,;\;
\emph{Model}: \texttt{DeepSeek-V3}\,;\;

\noindent\textbf{Model-provided reason.}
``The agent's experimental strategy lacks control after the initial crosses. Instead of designing specific crosses to test hypotheses (e.g., F1 intercrosses, backcrosses), it defaults to a simplistic plan of crossing the next available sequential organisms (e.g., 4x5, then 6x7). This approach does not effectively isolate variables and leads to confounded results."

\begin{tcolorbox}[breakable]
\colorbox{gray!70}{%
    \parbox{\dimexpr\textwidth-2\fboxsep}{%
    \centering
        \textbf{Message 18}%
    }%
}
\textbf{\textless think\textgreater} \\
\textit{
\#\#\# Execution Results:\\- The note summarizing the cross results and next steps has been successfully added.\\\\\#\#\# Next Steps:\\- Proceed with the cross between Organism 4 (ID: 4) and Organism 5 (ID: 5) to study trait interactions and lethality among offspring.
}\\
\textbf{\textless /think\textgreater}\\
\textbf{\textless tool\_call\textgreater}
\begin{jsoncode}
{
  "id": "call_oamyze2hr99vw7004wwlmu7y",
  "type": "function",
  "function": {
    "name": "conduct_cross",
    "arguments": "{\"parent1_id\":4,\"parent2_id\":5,\"num_offspring\":10}"
  }
}
\end{jsoncode}
\textbf{\textless /tool\_call\textgreater}\\
\colorbox{gray!70}{%
    \parbox{\dimexpr\textwidth-2\fboxsep}{%
    \centering
        \textbf{Message 22}%
    }%
}
\textbf{\textless think\textgreater} \\
\textit{
\#\#\# Execution Results:\\- The note summarizing the cross results and next steps has been successfully added.\\\\\#\#\# Next Steps:\\- Proceed with the cross between Organism 6 (ID: 6) and Organism 7 (ID: 7) to further study trait interactions and lethality.
}\\
\textbf{\textless /think\textgreater}\\
\textbf{\textless tool\_call\textgreater}
\begin{jsoncode}
{
  "id": "call_a8d4t0xjqkf24ntwxgsqpydz",
  "type": "function",
  "function": {
    "name": "conduct_cross",
    "arguments": "{\"parent1_id\":6,\"parent2_id\":7,\"num_offspring\":10}"
  }
}
\end{jsoncode}
\textbf{\textless /tool\_call\textgreater}\\
\colorbox{gray!70}{%
    \parbox{\dimexpr\textwidth-2\fboxsep}{%
    \centering
        \textbf{Message 26}%
    }%
}
\textbf{\textless think\textgreater} \\
\textit{
\#\#\# Execution Results:\\- The note summarizing the cross results and next steps has been successfully added.\\\\\#\#\# Next Steps:\\- Proceed with the cross between Organism 8 (ID: 8) and Organism 9 (ID: 9) to further study trait interactions and lethality.
}\\
\textbf{\textless /think\textgreater}\\
\textbf{\textless tool\_call\textgreater}
\begin{jsoncode}
{
  "id": "call_de55l14gp0e56vk5u81l6kut",
  "type": "function",
  "function": {
    "name": "conduct_cross",
    "arguments": "{\"parent1_id\":8,\"parent2_id\":9,\"num_offspring\":10}"
  }
}
\end{jsoncode}
\textbf{\textless /tool\_call\textgreater}\\
\end{tcolorbox}

\subsubsection{Case Trajectory of Error Propagation \;(Root: I \& II)}
\noindent\textbf{Setup.} \emph{Environment}: \textit{Alien Genetics Laboratory}\,;\;
\emph{Model}: \texttt{GLM-4.5}\,;\;

\noindent\textbf{Model-provided reason.}
``The error of calling \texttt{conduct\_cross} when at maximum organism capacity propagates throughout the second half of the trace. The agent fails to change its strategy to prevent this error, repeating the same mistake and recovery cycle multiple times without learning from the negative feedback."

\begin{tcolorbox}[breakable]
\colorbox{gray!70}{%
    \parbox{\dimexpr\textwidth-2\fboxsep}{%
    \centering
        \textbf{Message 48}%
    }%
}
\textbf{\textless tool\_response\textgreater}
\begin{jsoncode}
{
  "role": "tool",
  "content": "Observed output of cmd `conduct_cross` executed:\n{'success': False, 'message': 'Laboratory organism capacity will be exceeded. Remove some organisms first.'}",
  "name": "conduct_cross",
  "tool_call_id": null,
  "index": 48
}
\end{jsoncode}
\textbf{\textless /tool\_response\textgreater}\\
\colorbox{gray!70}{%
    \parbox{\dimexpr\textwidth-2\fboxsep}{%
    \centering
        \textbf{Message 56}%
    }%
}
\textbf{\textless tool\_response\textgreater}
\begin{jsoncode}
{
  "role": "tool",
  "content": "Observed output of cmd `conduct_cross` executed:\n{'success': False, 'message': 'Laboratory organism capacity will be exceeded. Remove some organisms first.'}",
  "name": "conduct_cross",
  "tool_call_id": null,
  "index": 56
}
\end{jsoncode}
\textbf{\textless /tool\_response\textgreater}\\
\colorbox{gray!70}{%
    \parbox{\dimexpr\textwidth-2\fboxsep}{%
    \centering
        \textbf{Message 62}%
    }%
}
\textbf{\textless tool\_response\textgreater}
\begin{jsoncode}
{
  "role": "tool",
  "content": "Observed output of cmd `conduct_cross` executed:\n{'success': False, 'message': 'Laboratory organism capacity will be exceeded. Remove some organisms first.'}",
  "name": "conduct_cross",
  "tool_call_id": null,
  "index": 62
}
\end{jsoncode}
\textbf{\textless /tool\_response\textgreater}\\
\colorbox{gray!70}{%
    \parbox{\dimexpr\textwidth-2\fboxsep}{%
    \centering
        \textbf{Message 68}%
    }%
}
\textbf{\textless tool\_response\textgreater}
\begin{jsoncode}
{
  "role": "tool",
  "content": "Observed output of cmd `conduct_cross` executed:\n{'success': False, 'message': 'Laboratory organism capacity will be exceeded. Remove some organisms first.'}",
  "name": "conduct_cross",
  "tool_call_id": null,
  "index": 68
}
\end{jsoncode}
\textbf{\textless /tool\_response\textgreater}\\
\colorbox{gray!70}{%
    \parbox{\dimexpr\textwidth-2\fboxsep}{%
    \centering
        \textbf{Message 74}%
    }%
}
\textbf{\textless tool\_response\textgreater}
\begin{jsoncode}
{
  "role": "tool",
  "content": "Observed output of cmd `conduct_cross` executed:\n{'success': False, 'message': 'Laboratory organism capacity will be exceeded. Remove some organisms first.'}",
  "name": "conduct_cross",
  "tool_call_id": null,
  "index": 74
}
\end{jsoncode}
\textbf{\textless /tool\_response\textgreater}\\
\colorbox{gray!70}{%
    \parbox{\dimexpr\textwidth-2\fboxsep}{%
    \centering
        \textbf{Message 80}%
    }%
}
\textbf{\textless tool\_response\textgreater}
\begin{jsoncode}
{
  "role": "tool",
  "content": "Observed output of cmd `conduct_cross` executed:\n{'success': False, 'message': 'Laboratory organism capacity will be exceeded. Remove some organisms first.'}",
  "name": "conduct_cross",
  "tool_call_id": null,
  "index": 80
}
\end{jsoncode}
\textbf{\textless /tool\_response\textgreater}\\
\end{tcolorbox}

\subsubsection{Case Trajectory of Environment Mis-modeling \;(Root: II)}
\noindent\textbf{Setup.} \emph{Environment}: \textit{Sequence Exploration Environment}\,;\;
\emph{Model}: \texttt{Gemini-2.5-pro}\,;\;

\noindent\textbf{Model-provided reason.}
``The agent correctly deduces that the \texttt{step\_number} is a hidden state affecting the rules, but it incorrectly models its effect on Rule 1 as a simple odd/even switch. It observes a persistent gap between its predictions based on this model and the actual outputs but fails to update its internal representation of the rule, clinging to the flawed model."

\begin{tcolorbox}[breakable]
\colorbox{gray!70}{%
    \parbox{\dimexpr\textwidth-2\fboxsep}{%
    \centering
        \textbf{Message 42}%
    }%
}
\textbf{\textless think\textgreater} \\

\textbf{\textless /think\textgreater}\\
\textbf{\textless tool\_call\textgreater}
\begin{jsoncode}
{
  "id": "call_d21f070d151b4d72a304c98bf2c0580c",
  "type": "function",
  "function": {
    "name": "note_tool",
    "arguments": "{\"action\":\"write_note\",\"note\":\"\\n## Game Notes (After Run 10):\\n\\n### Rule 1: Combination\\n- **Runs 9 \\u0026 10 `(\\\"ABABA\\\", \\\"BABAB\\\")`:**\\n    - **Run 9 (step 9, odd):** `ABBAABBAAB`\\n    - **Run 10 (step 10, even):** `BAABBAABBA`\\n- **Observation:** The `step_number`'s parity (odd/even) seems to be the key.\\n- **Hypothesis:**\\n    - If `step_number` is odd, the rule is `main[i] + vice[i]`.\\n    - If `step_number` is even, the rule is `vice[i] + main[i]`.\\n- **Let's check this against all previous runs:**\\n    - Step 1 (odd): `(\\\"ABCDE\\\", \\\"BCDEA\\\")` -\\u003e `ABBCCDDEEA` (`main+vice`). Correct.\\n    - Step 2 (even): `(\\\"AABBC\\\", \\\"BCCDD\\\")` -\\u003e `BACACBDBDC` (`vice+main`). Correct.\\n    - Step 3 (odd): `(\\\"ACEDB\\\", \\\"ABCDE\\\")` -\\u003e `AACBECDDDE` (`main+vice`). Correct.\\n    - Step 4 (even): `(\\\"AABCD\\\", \\\"EEDCB\\\")` -\\u003e `EAEADBCCBD` (`vice+main`). Correct.\\n    - Step 5 (odd): `(\\\"AAAAB\\\", \\\"BBBBA\\\")` -\\u003e `ABABABABBA`. `main+vice` would be `A+B`, `A+B`, `A+B`, `A+B`, `B+A`. This is `ABABABABBA`. Correct.\\n    - Step 6 (even): `(\\\"ABCDE\\\", \\\"BCCDD\\\")` -\\u003e `BACBCCDDDE` (`vice+main`). Correct.\\n    - Step 7 (odd): `(\\\"ABCDE\\\", \\\"AABBC\\\")` -\\u003e `AABACBDBEC`. `main+vice` would be `A+A`, `B+A`, `C+B`, `D+B`, `E+C`. This is `AABACBDBC`. Close, but not exact. Let's re-check the output: `AABACBDBEC`. My manual concatenation was wrong. `A+A`, `B+A`, `C+B`, `D+B`, `E+C` -\\u003e `AABACBDBC`. The actual output is `AABACBDBEC`. There is a `D` and `E` swapped.\\n    - Step 8 (even): `(\\\"ABCDE\\\", \\\"AABBC\\\")` -\\u003e `AAABBCBDCE`. `vice+main` would be `A+A`, `A+B`, `B+B`, `B+C`, `C+D`. `AABB BCD`. The actual is `AAABBCBDCE`.\\n- **Conclusion:** The odd/even `step_number` hypothesis is very strong, but there are still some inconsistencies. It's the best lead I have for Rule 1.\\n\\n### Rule 2: Substitution and Palindrome\\n- **Runs 9 \\u0026 10 `(\\\"ABABA\\\", \\\"BABAB\\\")`:**\\n    - **Run 9:** `ABBAABBAAB` -\\u003e `KJJKKJJKKJJKKJJKKJJK`\\n    - **Run 10:** `BAABBAABBA` -\\u003e `KLLKKLLKKLLKKLLKKLLK`\\n- **Observation:**\\n    - In Run 9, A-\\u003eK, B-\\u003eJ.\\n    - In Run 10, A-\\u003eL, B-\\u003eK.\\n- **Hypothesis:** The substitution cipher changes with each step.\\n    - Step 9: A-\\u003eK, B-\\u003eJ\\n    - Step 10: A-\\u003eL, B-\\u003eK\\n- **Conclusion:** The substitution is not fixed. It's dynamic, likely changing based on the `step_number`. This makes it extremely difficult to crack without more data.\\n\\n### Rule 3: Appending\\n- **Hypothesis:** The number of appended characters is the `step_number`.\\n- **Run 9 (step 9):** Appended 9 'K's.\\n- **Run 10 (step 10):** Appended 10 'L's.\\n- **Conclusion:** This hypothesis is confirmed. The appended character seems to be the substitution of 'A' from the current step's cipher.\\n    - In Run 9, 'A' maps to 'K', and 'K' is appended.\\n    - In Run 10, 'A' maps to 'L', and 'L' is appended.\\n- **Revised Rule 3 Hypothesis:** Append `step_number` of characters, where the character is the result of substituting 'A' in the current step's substitution cipher.\\n\\n### Rule 4 \\u0026 5: Conditional Substitution\\n- **Runs 9 \\u0026 10 `(\\\"ABABA\\\", \\\"BABAB\\\")`:**\\n    - **Run 9:** Rule 4 active, Rule 5 inactive.\\n    - **Run 10:** Rule 4 active, Rule 5 inactive.\\n- **Observation:** The condition for Rule 5 being inactive might be related to the input sequences having only two unique characters.\\n- **Conclusion:** These rules are still not fully understood, but their behavior is likely also tied to the `step_number`.\\n\\n### Plan for Next Step:\\nI have a very strong hypothesis for Rule 3, and a decent one for Rule 1. Rule 2, 4, and 5 are still very difficult due to the dynamic substitution. I need to continue testing to gather more data on the substitution cipher.\\n\\nI will use a simple input with all unique characters in both sequences to see how the substitution behaves. This will give me a full mapping for A, B, C, D, E for a specific step.\\n\\n- **Input for Run 11:** `(\\\"ABCDE\\\", \\\"ABCDE\\\")`\\n- **Prediction for Rule 1 (step 11, odd):** `main+vice` -\\u003e `AABBCCDDEE`\\n- **Prediction for Rule 3:** Appends 11 characters, which will be the substitution for 'A'.\\n- **Goal:** Get a clear substitution mapping for A, B, C, D, E for step 11. This will be a key piece of data for cracking Rule 2.\\n\"}"
  }
}
\end{jsoncode}
\textbf{\textless /tool\_call\textgreater}\\
\colorbox{gray!70}{%
    \parbox{\dimexpr\textwidth-2\fboxsep}{%
    \centering
        \textbf{Message 60}%
    }%
}
\textbf{\textless think\textgreater} \\

\textbf{\textless /think\textgreater} \\
\textbf{\textless tool\_call\textgreater}
\begin{jsoncode}
{
  "id": "call_a7768b6351d548c28d20f854f1b028fc",
  "type": "function",
  "function": {
    "name": "note_tool",
    "arguments": "{\"action\":\"write_note\",\"note\":\"\\n## Game Notes (After Run 13):\\n\\n### Rule 1: Combination\\n- **Hypothesis:** `main+vice` for odd steps, `vice+main` for even steps.\\n- **Run 13 (step 13, odd) `(\\\"ABCDE\\\", \\\"AABBC\\\")`:**\\n    - **Prediction:** `main+vice` -\\u003e `A+A`, `B+A`, `C+B`, `D+B`, `E+C` -\\u003e `AABACBDBC`.\\n    - **Actual:** `AABACBDBEC`. Still a slight inconsistency.\\n- **Conclusion:** The odd/even hypothesis is very close, but there's a minor detail I'm missing. It's consistently off by one or two characters in some cases.\\n\\n### Rule 2: Substitution and Palindrome\\n- **Run 13 (step 13):** `AABACBDBEC` -\\u003e `PROQOPNONNNNONPOQORP`.\\n- **Substitution Mapping for Step 13:**\\n    - A -\\u003e P, B -\\u003e R, C -\\u003e O, D -\\u003e Q, E -\\u003e N.\\n- **Hypothesis:** `char -\\u003e alphabet[step_number + 2 - alphabet.index(char)]`\\n    - `char -\\u003e alphabet[15 - index]`\\n    - A -\\u003e P (15). Correct.\\n    - B -\\u003e O (14). Actual is R.\\n- **Conclusion:** The substitution is not a simple formula. It's the most difficult part of the puzzle.\\n\\n### Rule 3: Appending\\n- **Run 13 (step 13) `(\\\"ABCDE\\\", \\\"AABBC\\\")`:** Appended 3 'O's.\\n- **Data for this input:**\\n    - Step 7: 7 appended.\\n    - Step 8: 8 appended.\\n    - Step 13: 3 appended.\\n- **Observation:** The number of appended characters is not `step_number`. It's not `step_number mod X`. It's not related to the number of unique characters in a simple way.\\n- **New Hypothesis:** The number of appended characters is `(sum of alphabet positions of main_sequence) mod (step_number)`.\\n    - Run 13: `main` is \\\"ABCDE\\\". Sum = 0+1+2+3+4 = 10. `10 mod 13 = 10`. Actual is 3.\\n- **Hypothesis 2:** `(sum of alphabet positions of vice_sequence) mod (step_number)`.\\n    - Run 13: `vice` is \\\"AABBC\\\". Sum = 0+0+1+1+2 = 4. `4 mod 13 = 4`. Actual is 3.\\n- **Conclusion:** Rule 3 is still unsolved.\\n\\n### Rule 4 \\u0026 5: Conditional Substitution\\n- **Run 13:** Both rules are active.\\n- **Conclusion:** I am still far from understanding these rules.\\n\\n### Overall Progress and Plan\\nI have a good understanding of Rule 1 (mostly). Rule 2's palindrome is clear, but the substitution is very complex. Rule 3's appended character is known, but the number is not. Rules 4 and 5 are still black boxes.\\n\\nGiven the complexity and the number of steps remaining, I will switch my strategy. Instead of trying to find a universal formula for each rule, I will try to find patterns in the outputs of Rules 4 and 5. I will use very simple inputs to see if I can force a simple output from these rules.\\n\\n- **Input for Run 14:** `(\\\"AAAAE\\\", \\\"BBBBC\\\")`\\n- **Goal:** This input is very low in complexity. I want to see how Rules 4 and 5 behave. I expect a simple output that might reveal their logic.\\n\"}"
  }
}
\end{jsoncode}
\textbf{\textless /tool\_call\textgreater}\\
\end{tcolorbox}


\section{Disclosure of LLM Usage}
\label{sec:a6}

The LLM was exclusively used during the editing (e.g., grammar, spelling, word choice). It plays no role in the ideation, research methodology, experimental design, or data analysis. All research concepts, ideas, and analyses were independently developed and conducted by the authors. The authors are fully accountable for the manuscript, including any text generated or refined by the LLM, to ensure compliance with ethical guidelines and prevent plagiarism or scientific misconduct.

\end{document}